\documentclass[10pt,journal,compsoc]{IEEEtran}
\usepackage{mathptmx}  
\usepackage{ragged2e} 

\usepackage{times}
\usepackage{soul}
\usepackage{url}
\usepackage[hidelinks]{hyperref}
\usepackage[utf8]{inputenc}
\usepackage[caption=false]{subfig}
\makeatletter
\usepackage{float}
\usepackage{amsmath}
\usepackage{amsfonts}
\usepackage{booktabs}

\usepackage{amssymb, amsthm}
\usepackage{algorithm}
\usepackage{algorithmic}
\usepackage{pgfplots}
\pgfplotsset{compat=1.7}
\usepackage{tikz}
\usepackage{pifont}
\usepackage{xcolor}
\usepackage{placeins}
\usepackage{booktabs}
\usepackage{multirow}
\usepackage{wasysym}
\usepackage{longtable}
\usepackage{setspace}
\usepackage{comment}
\theoremstyle{plain}
\newtheorem{assumption}{Assumption}
\newtheorem{theorem}{Theorem}[section]

\newtheorem{lemma}[theorem]{Lemma}

\definecolor{ao(english)}{rgb}{0.0, 0.5, 0.0}
\definecolor{bronze}{rgb}{0.8, 0.5, 0.2}
\definecolor{deeppink}{rgb}{1.0, 0.08, 0.58}
\definecolor{sepia}{rgb}{0.44, 0.26, 0.08}
\definecolor{aqua}{rgb}{0.0, 1.0, 1.0}
\definecolor{ao(english)}{rgb}{0.0, 0.5, 0.0}
\definecolor{bronze}{rgb}{0.8, 0.5, 0.2}
\definecolor{deeppink}{rgb}{1.0, 0.08, 0.58}
\definecolor{sepia}{rgb}{0.44, 0.26, 0.08}
\definecolor{aqua}{rgb}{0.0, 1.0, 1.0}
\definecolor{amber}{rgb}{1.0, 0.75, 0.0}
\definecolor{darkmagenta}{rgb}{0.55, 0.0, 0.55}
\definecolor{falured}{rgb}{0.5, 0.09, 0.09}
\definecolor{brass}{rgb}{0.71, 0.65, 0.26}
\definecolor{coolblack}{rgb}{0.0, 0.18, 0.39}
\usepackage{silence}

\ifCLASSOPTIONcompsoc
  \usepackage[nocompress]{cite}
\else
  \usepackage{cite}
\fi

\ifCLASSINFOpdf
\else
\fi

\begin{document}

\title{Implicit Regularization for Multi-label Feature Selection}

\author{Dou~El~Kefel~Mansouri,~Khalid~Benabdeslem ~and~Seif-Eddine~Benkabou

}


\IEEEtitleabstractindextext{%
\begin{abstract}
\justifying
  In this paper, we address the problem of feature selection in the context of  multi-label learning, by using a new estimator based on implicit regularization and label embedding. Unlike the sparse feature selection methods that use a penalized estimator with explicit regularization terms such as $l_{2,1}$-norm, MCP or SCAD, we propose a simple alternative method via Hadamard product parameterization. In order to guide the feature selection process, a latent semantic of multi-label information method is adopted, as a label embedding. Experimental results on some known benchmark datasets suggest that the proposed estimator suffers much less from extra bias, and may lead to benign overfitting. 
\end{abstract}

}

\maketitle
\IEEEdisplaynontitleabstractindextext
\IEEEpeerreviewmaketitle

\section{Introduction} 
\label{sec:introduction}
Multi-label learning focuses on the problem that each instance is associated with multiple class labels simultaneously \cite{gibaja2015tutorial,feng2019collaboration,panos2021large}, which is ubiquitous in many real-world applications, such as image annotation \cite{liu2018markov,liu2018svm,kundu2020exploiting}, text categorization \cite{liu2017deep,lee2019memetic}, and gene function classification \cite{fodeh2018exploiting,huang2020identification}. Similar to single-label learning, high-dimensional data with an enormous amount of redundant features significantly increases the computational burden of multi-label learning, which could also lead to over-fitting and performance degradation of learning algorithms \cite{wu2020multi}. 

To deal with this problem, feature selection represents a very effective way to alleviate the curse of dimensionality by selecting the most informative feature subsets from the original set. Many multi-label feature selection algorithms have been proposed to find a lower-dimensional representation of the original feature space, which can be broadly classified into transformation based methods that transform the multi-label data into either one or more single-label data subsets \cite{lapin2017analysis}, \cite{ma2018topic}, and direct methods that adapt the popular learning techniques to multi-label setting, without requiring any preprocessing \cite{huang2017joint}, \cite{pereira2018categorizing}, \cite{zhang2020multi}. In the first category, feature selection approaches can be used directly as filter like Fisher score \cite{hart2000pattern}, wrapper like sequential feature selection \cite{kohavi1997wrappers} or embedded like Lasso \cite{tibshirani1996regression}. In the second category, the feature selection approaches are revised in order to handle the multi-label parameter. We note in this category, the Robust Feature Selection (RFS) based on $l_ {2,1}$-norm regularization \cite{nie2010efficient}, the Multi-label dimensionality reduction method (MDDM) \cite{zhang2010multilabel}, the Multi-label Informed Feature Selection (MIFS) \cite{jian2016multi}, the Multi-label feature selection algorithm based on ant Colony Optimization (MLACO) \cite{paniri2020mlaco}, the ensemble method for semi-supervised multi-label feature selection (3-3FS) \cite{alalga20213} and  the Global and Local Feature Selection (GLFS) \cite{zhang2022integrating}, not to mention more. 

Particularly, sparse feature selection methods in the context of multi-label learning had a great deal of attention in recent years. Thanks to their estimators with explicit regularization schemes, like $l_{2,1}$-norm \cite{nie2010efficient}, $l_{2,1/2}$-norm \cite{shi2015sparse}, MCP \cite{shi2019feature} or SCAD \cite{fan2001variable}, the effect of the curse of dimensionality has been greatly mitigated and the learning process has been enormously improved. 
\\

\textbf{Motivation and our contribution.}
Despite the success of multi-label-learning-based  sparse feature selection methods, they suffer from \textit{extra bias} due to the regularization term introduced artificially to restrict the effective size of the parameter space \cite{vito2005learning,yao2007early}. Somewhat more clearly, adding some kind of norm constraint to an objective function of interest makes the modified optimization problem more complex. Hence,  the overall estimator may be deteriorated and may not fall below the penalty level to accommodate a possibly faster convergence rate. 

The aim of this paper is to propose a new estimator for multi-label feature selection, which suffers less bias than usual explicitly penalized estimators, and leads to more regular solutions. 
The proposed estimator is mainly based on an \textit{implicit regularizer} used to prevent the overfitting. The implicit regularizer leads to a change-of-variable via a simple \textit{Hadamard product parametrization} (element-wise product) \cite{hoff2017lasso}. The latter  plays the role of explicit penalty but  by transforming the penalized multi-label-learning-based sparse feature selection problem into an unconstrained smooth problem and also provides numerical stability (detail in section \ref{method}). Even if the implicit regularization has received particular attention in recent years, it has been addressed most often in the context of learning \cite{li2022benign}, \cite{zhou2023implicit}, \cite{shamir2023implicit}.
 In a different direction, this paper considers implicit regularization outside the learning process (see section \ref{method}). 

Furthermore, we used the \textit{latent semantic analysis} in conjunction with  the implicit regularization  to guide the feature selection process. This choice is due to the fact that estimating the correlation between features and class labels is often difficult due to the presence of noise and unnecessary information labels in the label set. Thus, motivated by Latent Semantic Indexing (LSI) \cite{deerwester1990indexing,landauer1998introduction,dumais2004latent}, it is possible to decompose the multi-labeled output space into a small dimension space cleaned of noise and unnecessary labels, and use this low-dimensional space to guide the feature selection process, via an implicit regularization. 

Interestingly, our estimator gives rise to the \textit{benign overfitting} phenomenon that typically occurs when the training error is significantly smaller than the test error, yet still maintains strong generalization performance. Although the explanation for the cause of this phenomenon remains a mystery to researchers, they are almost inclined to believe that implicit regularization is one of its mechanisms \cite{li2022benign}, \cite{shamir2023implicit}. They also confirm that explicit regularization in the form $l_1$ or $l_2$ does not lead to its emergence \cite{zhou2023implicit}. 
Thus, we experimentally show, through this paper, that another implicit regularization effect independent of the learning process could also lead to the emergence of this phenomenon. 

To the best of our knowledge, this is the first work considering together  a label embedding and an implicit regularization  in the context of the multi-label feature selection while  overcoming the extra bias. The main contributions of this paper with respect to the relative literature are summarized as follows:
\begin{enumerate}
	\item We propose a novel framework for multi-label feature selection, named mFSIR, i.e. \textbf{m}ulti-label \textbf{F}eature \textbf{S}election \textbf{I}mplicit \textbf{R}egularization.
	\item The proposed framework is based on a new estimator  that overcomes the extra bias.
	\item The proposed estimator relies on implicit regularization via Hadamard product parametrization in conjunction with the label embedding. 		
	\item We conduct experiments on some known benchmark  datasets to validate our proposal with different scenarios. 
\end{enumerate}

\textbf{Outline.} The rest of the paper is organized as follows. Section \ref{related_work} reviews some related works on machine learning using the explicit and implicit regularization terms as well as the latent semantics analysis. Section \ref{method} describes our proposed approach. Section \ref{experiment} provides some experimental results for validating the approach on some known datasets. Section \ref{conclusion} draws conclusion and some future directions from this work.
\section{Related work}
\label{related_work}
In this section, we first present the explicit and the implicit regularization concepts; then, we present  the concept of latent semantics of multi-label information and a brief overview of multi-label data selection approaches.

\subsection{Explicit/implicit regularization}
The literature on sparse feature selection methods in the context of multi-label learning is extensive. Jian \textit{et al} \cite{jian2016multi,jian2018exploiting}. proposed a novel multi-label informed feature selection framework called MIFS, that exploits label correlations to select discriminative features across multiple labels. He \textit{et al}. \cite{he2019joint} introduced a multi-label classification approach joint with  label correlations, missing labels and feature selection. Hu \textit{et al}. \cite{hu2020robust} proposed a robust multi-label feature selection with dual-graph regularization. Zhang \textit{et al}. \cite{zhang2020multi} suggested a new method that exploits both view relations and label correlations to select discriminative features for further learning. Fan \textit{et al}. \cite{fan2021manifold} proposed a manifold learning with structured subspace for multi-label feature selection. Huang \textit{et al}. \cite{huang2021multi} introduced a Multi-label feature selection via manifold regularization and dependence maximization.

All the above-mentioned works connect an  explicit regularization term to the gradient descent optimization in order to mitigate the curse of dimensionality and improve the learning process. Nevertheless, the explicit regularization is not sufficient for controlling the generalization error \cite{zhang2021understanding}, and may lead to a less accurate estimation due to the large bias \cite{rakitianskaia2015measuring}. In this context, some recent work claim that regularization may also be implicit, and the generalization error may enormously improved using implicit regularization than explicit one. For instance, Yu \textit{et al}. used the early stopping as an implicit regularization  to improve  prediction. Zhang \textit{et al}. \cite{zhang2021understanding} conducted a study to understand deep learning. They showed that explicit forms of regularization do not adequately explain the generalization error and the neural networks generalize well even without explicit regularization.  Vaskevicius \textit{et al}. \cite{vaskevicius2019implicit} proposed an algorithm based on implicit regularization for sparse linear regression. They showed that, unlike explicit regularization, algorithms based on implicit regularization applied to a sparse recovery problem adapt to the problem difficulty and yield optimal statistical rates. Zhao \textit{et al}. \cite{zhao2019implicit} considered implicit regularization for solving least square problems in the context of linear regression. They illustrated advantages of using implicit regularization via gradient descent over parametrization in sparse vector estimation. Recently, Li \textit{et al}. \cite{li2022benign}. proved that the implicit regularization could exhibit, under certain conditions, the phenomenon of benign overfitting. This was confirmed later in \cite{shamir2023implicit}, where the authors also provided the situations in which benign overfitting can occur. Chatterji \textit{et al}. \cite{chatterji2022interplay} argued that the implicit regularization is essential  in determining the generalization properties of the learnt model. Zhou \textit{et al} \cite{zhou2023implicit}. affirmed that one of the major explanations for benign overfitting is implicit regularization.

\subsection{Latent semantics of multi-label information} 
Furthermore, Latent Semantics Analysis (LSA) was originally developed, and has been most commonly applied to, for improving information retrieval  \cite{dumais2004latent}. This by using dimensionality reduction techniques that preserves the information of inputs and meanwhile captures the correlations between the multiple outputs. Yu \textit{et al}. \cite{yu2005multi} proposed a Multi-label informed Latent Semantic Indexing (MLSI) that maintains the inputs and simultaneously captures correlations between multiple output. Changqing \textit{et al}. \cite{zhang2018latent}  proposed a Latent Semantic Aware Multi-view Multi-label Learning (LSA-MML) that simultaneously seeks a predictive common representation of multiple views and the corresponding projection model between the common representation and labels. Zhang \textit{et al}. \cite{zhang2020multi} introduced a technique that projects the multi-labeled information into a reduced space by using the idea of latent semantic analysis.

\subsection{Multi-label feature selection} 
Multi-label feature selection approaches can be roughly split into problem transformation approaches and adaption approaches. The first category includes filters, wrappers and embedded approaches that transform the multi-label data into either one or more single-label data subsets. Most popular filter approaches are, Fisher score \cite{hart2000pattern}, ReliefF \cite{kononenko1994estimating} and f-statistic \cite{kong2012multi}. Most popular wrapper
approaches are, sequential feature selection \cite{skalak1994prototype}, randomized
feature selection \cite{duan2005multiple}, support vector machines and recursive feature elimination \cite{duan2005multiple}. The commonly used embedded approaches are Lasso \cite{tibshirani1996regression}, LARS \cite{efron2004least}, VS-CCPSO \cite{song2020variable} and NLE-SLFS \cite{zhang2019nonnegative}. The second category includes approaches that adapt the popular learning
techniques to multi-label setting, without requiring any pre-processing. Popular approaches in this category include: Robust Feature Selection (RFS) based on $l_ {2,1}$-norm regularization \cite{nie2010efficient}, Multi-label dimensionality reduction method (MDDM) \cite{zhang2010multilabel}, and Multi-label Informed Feature Selection (MIFS) \cite{jian2016multi}. 

  

\section{The mFSIR approach}
\label{method}
In this section, we first present the notations used throughout the paper and then introduce the formulation of our proposed method mFSIR. 

\subsection{Notations}
We use bold-faced symbols to denote vectors and matrices. Let $\mathbf{X}= [ \, \mathbf{x}_1,\mathbf{x}_2,...,\mathbf{x}_n ] \, \in \mathbb R_{}^{n \times m}$ be the instance matrix and $\mathbf{Y}= [ \, \mathbf{y}_1,\mathbf{y}_2,...,\mathbf{y}_n ] \, \in \mathbb \{0,1\}^{n \times q}$ be the label matrix. \textit{m} represents the size of feature vectors and \textit{q} represents the number of class labels $\{c_1,c_2,...,c_q\}$. $\mathbf{y}_i= [ \, y_{i1},y_{i2},...,y_{iq} ] \,\in \mathbb \{0,1\}^{q}$ is a binary vector, where $y_{ij}=1$, if $\mathbf{x}_{i}$ is associated with the label $c_{j}$ and $y_{ij}=0$, otherwise. We use $\left\|\mathbf{.}\right\|_{p}$ for the $l_p$-norm. The Frobenius norm ($l_{2,2}$) is defined as:
\begin{equation}
\left\|\mathbf{X}\right\|_{F} =\left( \sum_{i=1}^{m}\left\|\mathbf{X}_{i} \right\|_{2}^{2}\right) = \left(\sum_{i=1}^{m} \left(\sum_{j=1}^{n}\mathbf{X}_{ij}^{2}\right)\right)_{}^{1/2}
\end{equation}

\subsection{Problem statement}
Consider the usual sparse multi-label feature selection based on
an explicit regularization term \cite{nie2010efficient}:
\begin{equation}
\label{eqrfs}
\begin{split}
\textbf{$\Xi$}: \min_{\mathbf{W}}\left\|\mathbf{X}(\mathbf{W})-\mathbf{Y}\right\|^{2}_{F}+\underbrace{\gamma \left\|\mathbf{W}\right\|^{}_{2,1}}_{explicit\_term} 
\end{split}
\end{equation}
where, $\mathbf{W}\in \mathbb R_{}^{m \times q}$ is the feature coefficient matrix. $\gamma$ is a hyper-parameter used to control the strength of the regularization with respect to the loss function. The objective of \textbf{$\Xi$} is to make $\mathbf{W}$ well sparse, and this can be done by the second term based on the $l_{2,1}$ norm. The bigger  $\gamma$ is, the more the coefficients of $\mathbf{W}$ are reduced until they are exactly zero. 

In fact, Eq. (\ref{eqrfs}) is easy to be optimized and can have global optimal solutions. However, its explicit regularization term may expose it to large bias and thus lead it to less accurate estimation \cite{zhao2019implicit}.  Therefore, we assume that replacing the explicit regularizer with a simple implicit regularizer can overcome the extra bias. \subsection{Implicit regularization via  Hadamard product
parameterization}
Here, we rely on the works of \cite{biometrica} and  propose to  replace the explicit regularization in Eq. (\ref{eqrfs}) by an \textit{implicit} one  based on  the \textit{Hadamard product parametrization} (see Eq. (\ref{eqimplicit})).  
\begin{assumption}
\label{asump1}
Instead of using a direct coefficient matrix $\mathbf{W}$ to be regularized, we use the element-wise product of two matrices $\mathbf{G}$ $\in \mathbb R_{}^{m \times q}$ and  $\mathbf{H}$ $\in \mathbb R_{}^{m \times q}$ that should estimate $\mathbf{W}$.
\end{assumption}
From the Eq. (\ref{eqimplicit}), the \textit{term 2} in \textbf{$\Xi$} is used as an explicit regularization term.  Under the Assumption \ref{asump1}, the same term is dropped in \textbf{$\widehat\Xi$} and implicitly introduced in the form of ($\mathbf{G} \odot \mathbf{H}$).   
 \begin{equation}
 \label{eqimplicit}
\begin{split}
\textbf{$\Xi$}: \overbrace{\underbrace{\min_{\mathbf{W}}\left\|\mathbf{X}(\mathbf{W})-\mathbf{Y}\right\|^{2}_{F}}_{term 1}+\underbrace{\gamma \left\|\mathbf{W}\right\|^{}_{2,1}}_{term 2}}^{Explicit}  \hspace{0.5cm} \xrightarrow[\text{}]{\text{replaced by}} \hspace{0.5cm}\\
\textbf{$\widehat\Xi$ }: \overbrace{\underbrace{\min_{\mathbf{G,H}}\left\|\mathbf{X}(\mathbf{G} \odot \mathbf{H})-\mathbf{Y}\right\|^{2}_{F}}_{term 1}}^{Implicit}
\end{split}
\end{equation}
where, '$\odot$' is the Hadamard (element-wise) product.
\\\\
\begin{lemma}
\label{lem0}    
Based on \cite{hoff2017lasso}, a change-of-variable
via Hadamard product parametrization ($\mathbf{W} = \mathbf{G} \odot \mathbf{H}$), makes the non-smooth convex optimization problem for \textbf{$\Xi$} in Eq. (\ref{eqimplicit}) a smoothed optimization problem (\textbf{$\widehat\Xi$} in Eq. (\ref{eqimplicit})). 
\end{lemma}

Here, we should get \textbf{\^{G}} and \textbf{\^{H}}, the optimal values of $\mathbf{G}$ and $\mathbf{H}$. Thus, \textbf{\^{W}} = \textbf{\^{G}} $\odot$ \textbf{\^{H}} will represent the optimal value of $\mathbf{W}$. Details on obtaining \textbf{\^{G}} and \textbf{\^{H}} are in section \ref{optim}. We also take advantage of this parameterization  to create structured sparsity. The following assumption are made throughout the paper.
\begin{assumption}
\label{asump2}
Sparsity is ensured in \textbf{$\widehat\Xi$ } (Eq. (\ref{eqimplicit})) if the initial values of $\mathbf{G}$ or $\mathbf{H}$ are superior or equal to zero.
\end{assumption}
Since the objective of Eq. (\ref{eqrfs}) is to make $\mathbf{W}$ well sparse, we assume that this sparsity will take place if at least one of the matrices $\mathbf{G}$ or $\mathbf{H}$ is sparse due to the element wise product between the two matrices (see section \ref{Sparsit}). 

\subsection{Latent semantic analysis}
In multi-label learning, the correlation between features and class labels is often difficult to be estimated due to the presence of noise and unnecessary information labels in the label set. From this point, using latent semantics of multi-label information could be very effective to guide the feature selection process. 
\begin{assumption}
\label{asump3}
Based on \cite{dumais2004latent} and \cite{jian2016multi}, the multi-labeled output space $\mathbf{Y}$ can be decomposed to a product of two low-dimensional nonnegative matrices: $\mathbf{V} \in \mathbb R_{}^{n \times l}$ is the low-dimensional latent semantics matrix and $\mathbf{B} \in \mathbb R_{}^{l \times q}$ is the coefficient matrix of latent semantics, where $l<q$. 
\end{assumption}
Note that the nonnegative constraint is imposed on the decomposition phase since the latent semantic matrix obtained later will be more physically interpretable \cite{ding2006orthogonal}. 
Mathematically, the decomposition is done  by minimizing the following reconstruction error:
\begin{equation}
\label{equation3}
\min_{\mathbf{V\geq0,B\geq0}}\left\|\mathbf{Y}-\mathbf{VB}\right\|^{2}_{F}
\end{equation}
To ensure that local geometry structures are consistent between the input space $\mathbf{X}$ and the reduced low-dimensional semantics $\mathbf{V}$, it is important to add the following term.
\begin{equation}
\label{equation4}
\frac{1}{2}\sum_{i=1}^{n}\sum_{j=1}^{n}\mathbf{S}_{ij}(\mathbf{V}_{i:}-\mathbf{V}_{j:})^{2} =tr(\mathbf{V}^{T}(\mathbf{Z}-\mathbf{S})\mathbf{V})=tr(\mathbf{V}^{T}\mathbf{LV})
\end{equation}
where $\mathbf{S}_{ij}$ is the similarity matrix. $\mathbf{V}{_i}$ is the latent semantics of $y_i$. $\mathbf{Z}$ is a diagonal matrix with $\mathbf{Z}_{ii}$ = $\sum_{j=1}^{n}\mathbf{S}_{ij}$. $\mathbf{L} = \mathbf{Z} - \mathbf{S}$ is the graph laplacian matrix. The affinity graph $\mathbf{S}$ is modeled by Eq. (\ref{equation5}) \cite{cai2010unsupervised}, 
\begin{equation}
\label{equation5}
\mathbf{S}_{ij}= \begin{cases}            
e^{-\frac{\lVert \mathbf{x}_i-\mathbf{x}_j \rVert^2}{\lambda^2}} & \mbox{if} \hspace{0.2cm} \mathbf{x}_i \in N_p(\mathbf{x}_j) \hspace{0.2cm} or \hspace{0.2cm}  \mathbf{x}_j \in N_p(\mathbf{x}_i)\\
0 & otherwise,
\end{cases}
\end{equation}
where $N_p(\mathbf{x})$ denotes the \textit{p}-nearest neighbors of instance $\mathbf{x}$. By integrating the local geometric structure of the data, the Eq. (\ref{equation3}) becomes:
\begin{equation}
\label{equation6}
\min_{\mathbf{V\geq0,B\geq0}}\left\|\mathbf{Y}-\mathbf{VB}\right\|^{2}_{F}+\beta tr(\mathbf{V}^{T}\mathbf{LV})
\end{equation}
where $\beta$ is a regularization parameter, used to control local geometry structures.

\subsection{Objective function}
To perform feature selection, we take advantage of the low-dimensional latent semantics matrix $\mathbf{V}$ that encodes label correlations and greatly reduces the noise in the original multi-label output space. Hence, features most related to the latent semantics $\mathbf{V}$ will be chosen. Therefore, the objective function that we propose for multi-label feature selection and label decomposition, can be formulated as follows:
\begin{equation}
\begin{split}
\label{OF}
\min_{[\mathbf{G,H,V,B]\geq0}}\left\|\mathbf{X}(\mathbf{G}\odot\mathbf{H})-\mathbf{V}\right\|^{2}_{F}+\alpha \left\|\mathbf{Y}-\mathbf{VB}\right\|^{2}_{F}\\+\beta tr(\mathbf{V}^T\mathbf{L}\mathbf{V})
\end{split}
\end{equation}

The first term in Eq. (\ref{OF}) represents the \textit{primary contribution} to this paper, which is related to implicit regularization. The second and third terms contribute to the overall framework by enabling label embedding and improving feature selection. ($\mathbf{G} \odot \mathbf{H}$) determines a feature coefficient matrix  in which each row measures the importance of the $\textit{j}^{th}$ feature in approximating the latent semantics $\mathbf{V}$. $\alpha$ and $\beta$ are used to balance the second and the third terms in the equation Eq. (\ref{OF}), respectively.

\subsection{Optimization algorithm}
\label{optim}
Minimizing Eq. (\ref{OF}) jointly over $\mathbf{G}$, $\mathbf{H}$, $\mathbf{V}$ and $\mathbf{B}$ is a highly non-convex optimization problem with many saddle points, specially if the label space $\mathbf{Y}$ is noisy. 
Our objective function is differentiable at each variable, and its local minimizers can be found using an efficient alternating optimization algorithm. Thus, we can apply the gradient descent with Hadamard product
parameterization, by taking the derivative of the objective function w.r.t. variables $\mathbf{G}$, $\mathbf{H}$, $\mathbf{V}$ and $\mathbf{B}$, respectively:
\begin{equation}
\label{equation9}
\left\{
\begin{array}{l}
\mathbf{G}:=\mathbf{G}-\eta\nabla
f_{\mathbf{G}}(\mathbf{G},\mathbf{H},\mathbf{V},\mathbf{B})\\\\
\mathbf{H}:=\mathbf{H}-\eta\nabla f_{\mathbf{H}}(\mathbf{G},\mathbf{H},\mathbf{V},\mathbf{B})\\\\
\mathbf{V}:=P[\mathbf{V}-\eta\nabla f_{\mathbf{V}}(\mathbf{G},\mathbf{H},\mathbf{V},\mathbf{B})]\\\\
\mathbf{B}:=P[\mathbf{B}-\eta\nabla f_{\mathbf{B}}(\mathbf{G},\mathbf{H},\mathbf{V},\mathbf{B})]
\end{array}
\right.
\end{equation}
where
\begin{equation}
\begin{split}
\label{equation10}
\left\{
\begin{array}{l}
\nabla f_{\mathbf{G}}(\mathbf{G},\mathbf{H},\mathbf{V},\mathbf{B}):= \mathbf{H}\odot[2\mathbf{X}^{T}(\mathbf{X}(\mathbf{G}\odot \mathbf{H})-\mathbf{V})]\\\\
\nabla f_{\mathbf{H}}(\mathbf{G},\mathbf{H},\mathbf{V},\mathbf{B}):=\mathbf{G}\odot [2\mathbf{X}^{T}(\mathbf{X}(\mathbf{G} \odot \mathbf{H})-\mathbf{V})]\\\\
\nabla 
f_{\mathbf{V}}(\mathbf{G},\mathbf{H},\mathbf{V},\mathbf{B}):=2[(\mathbf{V}-\mathbf{X}(\mathbf{G}\odot\mathbf{H}))\\\hspace{3cm}
+\alpha(\mathbf{VB}-\mathbf{Y})\mathbf{B}^T+\beta\mathbf{LV}]\\\\
\nabla
f_{\mathbf{B}}(\mathbf{G},\mathbf{H},\mathbf{V},\mathbf{B}):=2[\alpha\mathbf{V}^T(\mathbf{VB}-\mathbf{Y})]
\end{array}
\right.
\end{split}
\end{equation}
and $P[{\mathbf{D}}]$ represents a box projection operator that maps the update $\mathbf{D}$ to a bounded region in order to ensure the
non-negativity:

\begin{equation}
\label{equation19}
P[{\mathbf{D}}]_{ij}= \begin{cases}            
\mathbf{D}_{ij} & \mbox{if} \hspace{0.2cm} \mathbf{D}_{ij} \geq 0\\
0 & otherwise,
\end{cases}
\end{equation}
 $\eta$ is a step size, used to accelerate the convergence
rate and to reduce the running time of the algorithm. Note that,  at each iteration,  one variable is updated while fixing the other three variables since the objective function is convex when any three variables are fixed. We illustrate the optimization of
Eq. (\ref{OF}) in Algorithm \ref{algorithm1}.
\begin{algorithm}
{\small
	\caption{mFSIR}
	\label{algorithm1}
	{\bfseries Input:} Data matrix
	$\mathbf{X}\in \mathbb R^{n\times m}$, Label matrix $\mathbf{Y}\in \mathbb R^{n\times q}$, Validation data: ($\overline{\mathbf{X}}, \overline{\mathbf{Y}}$), Parameters: $\alpha$, $\beta$, $\varpi$ $\eta$, $T_{max}$.
	
	{\bfseries Output:} Final estimate $(\textbf{\^{G}}\odot\textbf{\^{H}})$ and ranked features.
	\begin{algorithmic}[1] 
		\STATE 	Initialize
		
		 $[\mathbf{G}^{0}]\overset{\text{iid}}{\sim} Unif(-\varpi,\varpi),[\mathbf{H}^{0}]\overset{\text{iid}}{\sim}
		Unif(-\varpi,\varpi),[\mathbf{V}^{0}]\overset{\text{iid}}{\sim} Unif(-\varpi,\varpi), [\mathbf{B}^{0}]\overset{\text{iid}}{\sim} Unif(-\varpi,\varpi)$, iteration number $t=0$;
		\STATE \textbf{while} $t< T_{max}$ \textbf{do}\\
		\STATE \hspace{0.5cm}$\mathbf{G}^{t+1}:=\mathbf{G}^{t}-\eta \nabla
		f_{\mathbf{G}}(\mathbf{G},\mathbf{H},\mathbf{V},\mathbf{B})$;
		\\
		\STATE \hspace{0.5cm}$\mathbf{H}^{t+1}:=\mathbf{H}^{t}-\eta
		\nabla
		f_{\mathbf{H}}(\mathbf{G},\mathbf{H},\mathbf{V},\mathbf{B})$;\\
		\STATE 
		\hspace{0.4cm} $\mathbf{V}^{t+1}:= P[\mathbf{V}^{t}-\eta \nabla
		f_{\mathbf{V}}(\mathbf{G},\mathbf{H},\mathbf{V},\mathbf{B})]$;\\
		\STATE 
		\hspace{0.4cm} $\mathbf{B}^{t+1}:=P[\mathbf{B}^{t}-\eta \nabla
		f_{\mathbf{B}}(\mathbf{G},\mathbf{H},\mathbf{V},\mathbf{B})]$;\\
		\STATE \hspace{0.5cm} $t=t+1$\\
		\STATE \textbf{end}
	\end{algorithmic} }
\end{algorithm}

First, we apply  the updating formulas in Eq. (\ref{equation9}), with random initial values $\mathbf{G}^{0}, \mathbf{H}^{0}, \mathbf{V}^{0}$ and $\mathbf{B}^{0}$ chosen close enough to 0 in step 1. Notice that (0, 0, 0, 0) is a saddle point of the objective function, so we need to apply a small perturbation $\varpi$ on the initial values. We then apply the updating rules [lines 2 to 8] to get $\mathbf{G}, \mathbf{H}, \mathbf{V}$ and $\mathbf{B}$. Finally, the Hadamard product between the final values obtained from the two matrices $\mathbf{H}$ and $\mathbf{G}$ is estimated, and the resulting norm of each row is used to evaluate the relevance of the features to be selected.
\begin{lemma}
\label{lem2}
In the iterative process (Steps 2 to 8), $(\mathbf{G}^{t+1},\mathbf{H}^{t+1},\mathbf{V}^{t+1},\mathbf{B}^{t+1})$ tend to converge to a stationary point $(\mathbf{G}^{\infty},\mathbf{H}^{\infty},\mathbf{V}^{\infty},\mathbf{B}^{\infty})$ of Eq. (\ref{OF}) that satisfies the first order optimality condition :
\begin{equation*}
\left\{
\begin{array}{l}
\nabla f_{\mathbf{G}}(\mathbf{G}^{\infty},\mathbf{H}^{\infty},\mathbf{V}^{\infty},\mathbf{B}^{\infty})=0\\\\
\nabla f_{\mathbf{H}}(\mathbf{G}^{\infty},\mathbf{H}^{\infty},\mathbf{V}^{\infty},\mathbf{B}^{\infty})=0\\\\
\nabla f_{\mathbf{V}}(\mathbf{G}^{\infty},\mathbf{H}^{\infty},\mathbf{V}^{\infty},\mathbf{B}^{\infty})=0 \\\\
\nabla f_{\mathbf{B}}(\mathbf{G}^{\infty},\mathbf{H}^{\infty},\mathbf{V}^{\infty},\mathbf{B}^{\infty})=0
\end{array}
\right.
\end{equation*}
\end{lemma}
Stationary points of Eq. (\ref{OF}) can be local minimum, local maximum, or saddle points.  Under the Assumption \ref{asump3} that the label decomposition, by $\mathbf{VB}$ allows a good representation of the original label space $\mathbf{Y}$, we can consider that Eq. (\ref{OF}) does not have local maximum, all its local minimums are global minimum, and all saddle points are strict  \cite{zhao2019implicit}.
\begin{lemma}
\label{lem1}
Under the Assumption \ref{asump2}, Eq. (\ref{OF}) converges to a global minimum.
\end{lemma}
Eq. (\ref{eqrfs}) converges to a global minimum thanks to its external penalty  which forces certain coefficients of $\mathbf{W}$ to be null and thus reduces the dimensionality. More precisely, if $n>m$ (i.e, low-dimension regime) the equation admits a unique convex solution even without external penalty. In the case of $n<<m$ (i.e, high-dimension regime), the equation admits an infinity of solutions which means that regularization is so necessary to approximate the low-dimensional regime and thus approximate the unique solution.

Eq. (\ref{OF}) is subject to high-dimensional regime and the regularization, this time, is not explicit but rather implicit via Hadamard product parameterization. Under the assumptions \ref{asump1} and \ref{asump2}, 
one can approximate the low-dimensional regime. Therefore, the equation converges to a global minimum even with a simple gradient descent but provided that the step size $\eta$ is sufficiently small. Indeed, if $\eta$ is large, then the solution tends to move faster but at the risk of being local. 

\subsection{Time complexity analysis}

The computational complexity of
Algorithm \ref{algorithm1} is presented by the following
lemma.
\begin{lemma}
\label{lem3}
mFSIR is computed in time of $O(nml + nlq + n^2l)$    
\end{lemma}

mFSIR is computationally efficient because it only requires simple multiplication operations in the alternative optimization process. The calculation of the derivative w.r.t $\mathbf{G, H, V}$ and $\mathbf{B}$ is the major contributor to the computational complexity. Specifically, it takes $O(nml)$ to compute the derivatives w.r.t $\mathbf{G}$ and w.r.t $\mathbf{H}$. The cost for solving the derivative w.r.t $\mathbf{V}$ is $O(nml+nlq+n^2l)$. It needs $O(nlq)$ to calculate the derivative w.r.t $\mathbf{B}$. In our case, $l<<n$, $l<<m$ and $l<q$. Therefore, the overall cost for mFSIR is $O(nml + nlq + n^2l)$ in a single iteration.

\section{Experiments}
\label{experiment}
\textcolor{black}{In this section, we conduct a series of experiments on large datasets to validate the proposed framework. First, we evaluate the feature selection method on a toy dataset in terms of feature importance, visualization, and interpretability. Then, we perform an in-depth comparison with other multi-label feature selection algorithms, followed by further analysis.}

\subsection{Datasets and methods}
To validate the performance of mFSIR, we conduct experiments on one widely used dataset for feature selection: "Waveform" (available in UCI repository\footnote{https://archive.ics.uci.edu}), and ten multi-label benchmark datasets, available in MULAN Project\footnote{http://mulan.sourceforge.net/datasets.html}. Table \ref{table1} describes the characteristics of each multi-label dataset \textit{S}, including the number of examples ($\lvert S \rvert$), the number of features $(dim(S))$, the number of class labels $(L(S))$, the label cardinality $(LCard(S)$, and the label density $(LDen(S))$.  
\begin{table}[hbt!]
	\centering
    \caption{Multi-label datasets}%
    \label{table1}
    \scalebox{0.75}{
	\begin{tabular}{lcccccc}
		\hline 
		\textbf{Datasets} & $domain$ &$\lvert S \rvert$& $dim(S)$&$L(S)$&$LCard(S$&$LDen(S)$\\     
		\hline 
		\textbf{bibtex} &text &7395&1836&159&2.402&0.015\\	
		\textbf{Corel16k} &image &13770&500&153&2.859&0.019\\	
		\textbf{Delicious} &text (web)&16105&500&983&19.020&0.019\\
		\textbf{emotions} &music&593&72&6&1.869&0.311\\
		\textbf{Enron} &text &1702&1001&53&3.378&0.064\\
		\textbf{Language log}& text&1460 &1004&75&1.180&0.016\\
		\textbf{medical} &text &978&1449&45&1.245&0.028\\
		\textbf{scene} &image &2407&294&6&1.074&0.179\\	
		\textbf{tmc2007}& text&28596 &500&22&2.158&0.098\\
		\textbf{Yeast}& biology&2417 &103&14&4.237&0.303\\
		\hline
	\end{tabular}}
\end{table}

The performance of mFSIR is compared with several representative feature selection methods. A brief description of each method is given below.
\begin{itemize}
    \item \textbf{MIFS:} Multi-label Informed Feature Selection. It exploits the latent label correlations to select discriminative features across multiple labels \cite{jian2018exploiting}. [parameter configuration: $\alpha$, $\beta$, $\gamma$ $\in$ $\{10^{-4}, 10^{-3},...,10\}$];
   \item \textbf{MICO:} A mutual-information-based feature selection method, which obtains the optimal solution via constrained convex optimization with less time \cite{sun2019mutual}. [parameter configuration: $\alpha$, $\beta$ $\in$ $\{10^{-3}, 10^{-2},...,10^{3}\}$]; 
    \item \textbf{GRRO:} Multi-label Feature Selection via Global Relevance and Redundancy Optimization. A general global optimization framework, in which feature relevance, label relevance and feature redundancy are taken into account to facilitate the multi-label feature selection \cite{zhang2020multiGRRO}. [parameter configuration: $\alpha$, $\beta$ $\in$ $\{10^{-3}, 10^{-2},...,10^{3}\}$, \textit{k}=10]; 
    \textcolor{black}{\item \textbf{CMLL:} Compact Multi-Label Learning. It provides a compact learning that simultaneously integrates features and labels into a low-dimensional space to capture their dependencies and enhance performance in multi-label classification \cite{lv2021compact}. [parameter configuration: $m$ $\in$ $[5, 10, 20, 30]$, $\beta$, $\lambda$ $\in$ $\{10^{-3}, 10^{-2},...,10\}$];}
  \textcolor{black}{\item \textbf{LMLL:} Learning compact models for large-scale multi-label data. It retains the most discriminating label and feature parameters to produce compact models without sacrificing performance \cite{wei2019learning}. [parameter configuration: $\lambda \in [10^{-6}, 10^{-2}]$, $\delta \in [10, 500]$, $\epsilon \in [10^{-2}, 0.2]$] }
    \item \textbf{LassoNet:} A Neural Network with Feature Sparsity. It represents an extension of Lasso regression to  feed-forward neural networks with global feature selection \cite{lassonet}. [parameter configuration: lambda\_start=5e-1,
    path\_multiplier=1.05, \textit{M} = 10]. 
\end{itemize}

LassoNet uses early stopping as an implicit regularization and can be considered as direct competitor of mFSIR. Note that it has been adapted to multi-label selection since it is basically made for variable selection in single label.
\subsection{Experimantal Setup}
We adopt the  ML\textit{k}NN as a lazy classifier \cite{zhang2007ml}, where the parameter \textit{k} is set as 10. The perturbation parameter $\varpi$ is set as $10^{-5}$.
Five-fold cross validation is done to split training and test sets. The number of selected features is varied from 5\% to 30\% of the total number of features. 

For performance evaluation, we employ three metrics widely used  in multi-label learning for comparison, including: Hamming loss, Ranking loss 
and  Macro-averaged F1-score \cite{tsoumakas2007multi} \cite{xu2019robust} \cite{wu2017unified}. 
\begin{itemize}
    \item \textit{Hamming loss \cite{wu2017unified}\cite{alalga20213}:}
     $$\frac{1}{N}\sum_{i=1}^{N}\frac{ {\lVert h(x_i)\oplus y_i \rVert}_1}{C}$$
Here, $h(x_i)$ is the predicted label-vector, $y_i$ is the true label-vector, and $\oplus$ stands for the XOR operation between $h(x_i)$ and $y_i$.\\
    \item \textit{Ranking loss \cite{li2017improving}\cite{alalga20213}:}
      $$ \frac{1}{N}\sum_{i=1}^N\frac{|\{(y_s, y_t)\in Y_i \times \overline{Y_i}; \nonumber\\ f(x_i, y_s)\leq f(x_i, y_t)\}|}{|Y_i||\overline{Y_i}|}$$
Here, $\overline{Y_i}$ is the complementary set of $Y_i$ in $Y$.\\
\item \textit{Macro-averaged F1-score} \cite{sokolova2009systematic}:
\begin{equation*}
\frac{1}{M}\sum_{j=1}^{M}
\frac{
2\,\big|\{x \in a_j \mid \hat{y}_j(x)=1\}\big|
}{
|a_j| + \big|\{x \in b_j \mid \hat{y}_j(x)=1\}\big|
}
\end{equation*}
where $a_j=\{x_i \mid y_{ij}=1\}$ and $b_j=\{x_i \mid y_{ij}=0\}$.
\end{itemize}
For the first two evaluation metrics, the \textit{smaller} the metric value the better the performance. For the other evaluation metric, the \textit{larger} the metric value the better the performance.
\newpage

\subsection{\textcolor{black}{Visualization and interpretability}}
\begin{figure}[!ht]
		\centering
	\includegraphics[width=0.4\textwidth]{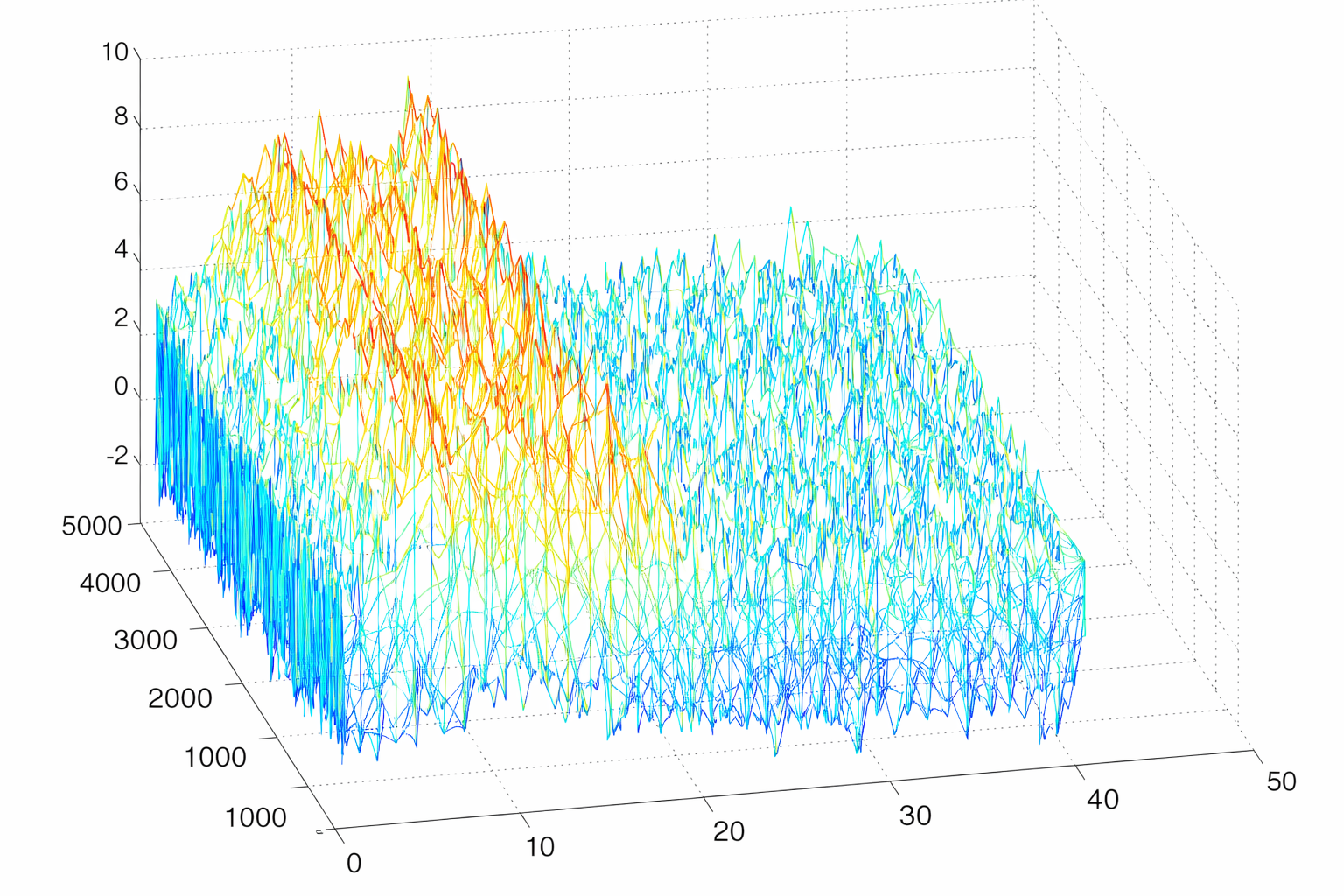}	
	\caption{Waveform dataset.}
	\label{fig:Wave}
\end{figure}

\begin{figure}[htbp]
\centering
\begin{minipage}[c]{0.40\textwidth}
    \centering
    \includegraphics[width=1.0\textwidth]{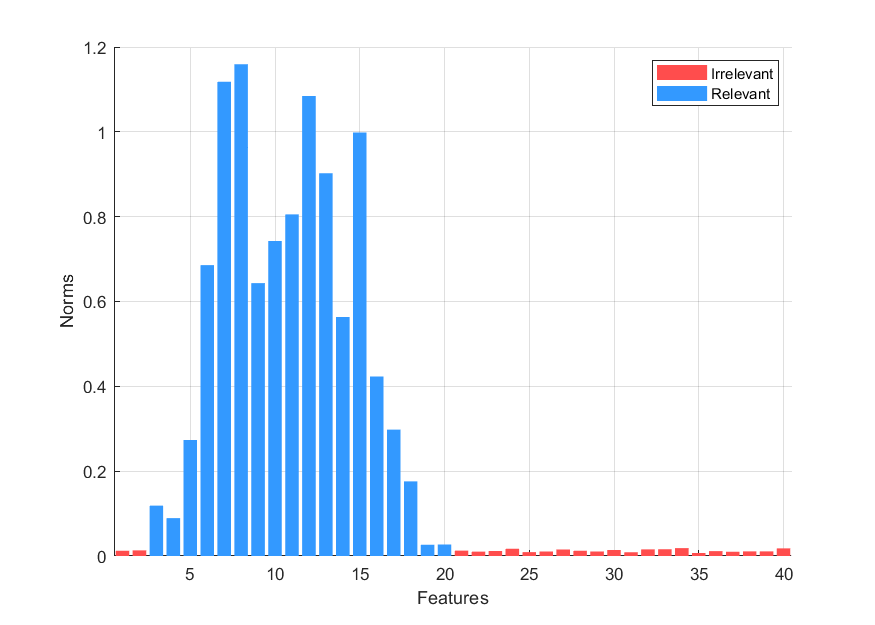}
    {\small (a)}
\end{minipage}
\begin{minipage}[c]{0.40\textwidth}
    \centering
    \includegraphics[width=1.0\textwidth]{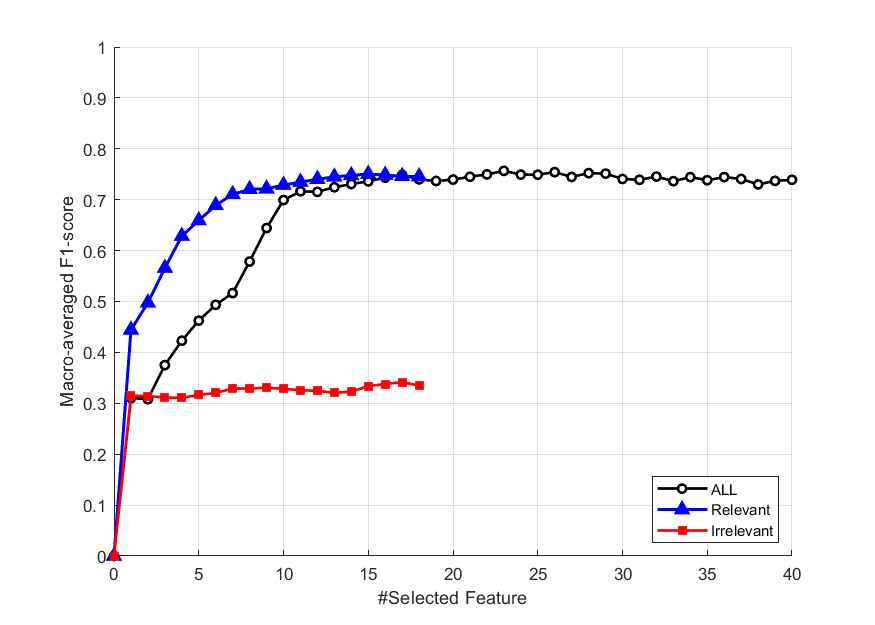}
    {\small (b)}
\end{minipage}
\begin{minipage}[c]{0.40\textwidth}
    \centering
    \includegraphics[width=1.0\textwidth]{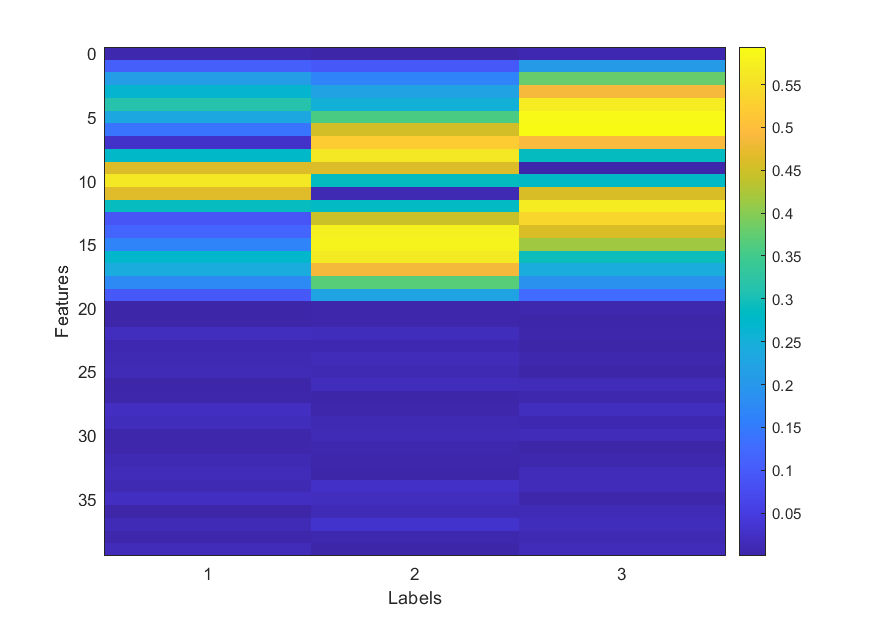}
    {\small (c)}
\end{minipage}
\begin{minipage}[c]{0.40\textwidth}
    \centering
    \includegraphics[width=1.0\textwidth]{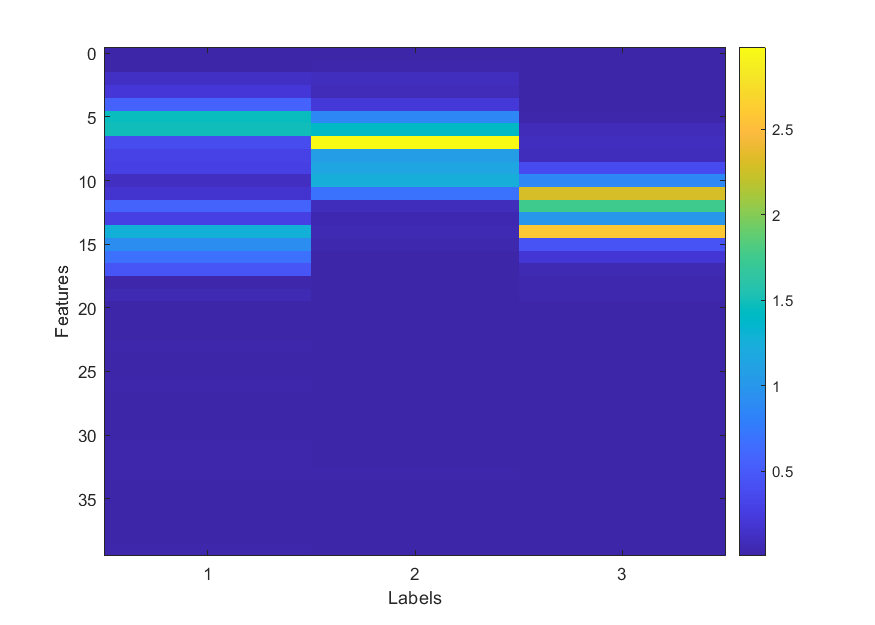}
    {\small (d)}
\end{minipage}

\caption{Interpretability of feature selection on "Waveform" dataset. 
(a) Feature importance,
(b) Feature quality,
(c) Feature-label correlations,
(d) Label-specific weighted importance.}
\label{fig:visualization}
\end{figure}

\textcolor{black}{In this section, we are particularly interested in visualizing the importance of features according to mFSIR outputs and their interpretability regarding the label space. For this analysis, we choose the "Waveform" datataset, which is widely used in machine learning and data mining tasks. This dataset consists of 5000 instances divided into 3 classes, and composed of 21 relevant features (the first ones) and 19 noise features with a mean of 0 and a variance of 1. Each class is generated from a combination of 2/3 ”base” waves (Fig. \ref{fig:Wave}).}
\textcolor{black}{First, the Waveform dataset, originally defined as a three-class multiclass problem, is converted into a three-dimensional binary label matrix via one-hot encoding in order to comply with the multi-output learning setting of mFSIR.\\
After applying mFSIR, we show different levels of visualization in Fig. \ref{fig:visualization}.}
\textcolor{black}{In Fig. \ref{fig:visualization}(a), we
present the importance of features provided by mFSIR according to their norms. Note that the higher the norm of $\left\|(\mathbf{G}\odot\mathbf{H})(j,:)\right\|$ the more relevant $j^{th}$ feature is. In this figure, we show the visualization of feature relevance with two colors. The red color represents the irrelevant features and the blue color represents the relevant ones. We
can see that the features (21 to 40) have low values on their norms, so the noise represented by these features is clearly detected.
In Fig. \ref{fig:visualization}(b), we show the classification performance in terms of Macro-averaged F1-score versus  the number of selected features with three curves. The black curve plots classification performance
using all the features (without ranking) in the dataset, while the red one represents classification performance with the irrelevant features detected by mFSIR. We can see that the performance remains very weak when the learning is undertaken using noise features only. The blue curve (plotted with relevant features) outperforms the black one and increases steadily over the first twenty features whose norms are the high ones on Fig. \ref{fig:visualization}(a). The Fig. \ref{fig:visualization}(c) shows the correlation between each label and all  features. We can see that there is no correlation with the noise (irrelevant features). The label 1 is strongly correlated with features \{9, 10, 11\}; the label 2 with \{6, 7, 8, 9, 13, 15, 16, 17\} and the label 3 with \{4, 5, 6, 7, 8, 11, 12, 13, 14\}. Finally, we present in Fig. \ref{fig:visualization}(d) the label specific weighted importance, which indicates the features that are effectively relevant with labels. Thus, the label 1 can be interpreted by the relevant features \{5, 6, 7, 13, 14\}; the label 2 by \{6, 7, 8, 9, 10\}; and the label 3 by \{10, 11, 12, 13, 14\}. That means that a feature can be irrelevant for a label even if the correlation between them is high. Thus, the interpretability of the label space is more related to selected features by mFSIR than a simple correlation between labels and features.}

\begin{table*}[!t]
	\centering
    \caption{Experimental results of comparing approaches (mean$\pm$std. deviation) on the ten datasets over three metrics.
($\downarrow$: the smaller the better; $\uparrow$: the larger the better).
The marker $\bullet$/$\circ$ indicates whether mFSIR is superior/inferior to the other method.
Win/tie/loss counts summarize pairwise comparisons between mFSIR and each comparing approach.
}
	\scalebox{0.7}{
	\begin{tabular}{cccccccccccp{2.7cm}}
				\hline
				\hline
	   & \multicolumn{11}{c} {\textbf{Hamming Loss} $\downarrow$}\\
				\cline{2-12}
    \multirow{-2}{*}{Dataset}	&bibtex&Corel16k&Delicious&emotions&Enron&language log&medical&scene&tmc2007&Yeast&\textbf{win/tie/loss counts for mFSIR}\\
		\hline
		MIFS&.011$\pm$.000$\circ$&.023$\pm$.000$\circ$&.019$\pm$.001$\bullet$&.341$\pm$.000$\bullet$&.234$\pm$.000$\bullet$&.020$\pm$.000$\circ$&.017$\pm$.000$\circ$&.172$\pm$.000$\bullet$&.137$\pm$.001$\bullet$&.205$\pm$.000$\circ$&\textbf{5/0/5}\\
   MICO&.018$\pm$.014$\bullet$&.052$\pm$.000$\bullet$&.022$\pm$.000$\bullet$&.269$\pm$.029$\bullet$&.360$\pm$.001$\bullet$&.289$\pm$.001$\bullet$&.033$\pm$.000$\circ$&.161$\pm$.006$\bullet$&.165$\pm$.000$\bullet$&.230$\pm$.011$\bullet$&\textbf{9/0/1}\\
   GRRO&.015$\pm$.000$\circ$&.021$\pm$.000$\circ$&.020$\pm$.000$\bullet$&.273$\pm$.000$\bullet$&.228$\pm$.000$\bullet$&.291$\pm$.000$\bullet$&.011$\pm$.000$\circ$&.245$\pm$.000$\bullet$&.103$\pm$.000$\bullet$&.312$\pm$.000$\bullet$&\textbf{7/0/3}\\

\textcolor{black}{CMLL}&\textcolor{black}{.462$\pm$.013$\bullet$}&\textcolor{black}{.439$\pm$.009$\bullet$}&\textcolor{black}{.499$\pm$.001$\bullet$}&\textcolor{black}{.330$\pm$.014$\bullet$}&\textcolor{black}{.455$\pm$.016$\bullet$}&\textcolor{black}{.330$\pm$.007$\bullet$}&\textcolor{black}{.563$\pm$.017$\bullet$}&\textcolor{black}{.325$\pm$.012$\bullet$}&\textcolor{black}{.381$\pm$.000$\bullet$}&\textcolor{black}{.451$\pm$.008$\bullet$}&\textcolor{black}{\textbf{10/0/0}}\\

\textcolor{black}{LMLL}&\textcolor{black}{.505$\pm$.003$\bullet$}&\textcolor{black}{.044$\pm$.001$\bullet$}&\textcolor{black}{.696$\pm$.000$\bullet$}&\textcolor{black}{.245$\pm$.010$\circ$}&\textcolor{black}{.464$\pm$.008$\bullet$}&\textcolor{black}{.407$\pm$.012$\bullet$}&\textcolor{black}{.510$\pm$.025$\bullet$}&\textcolor{black}{.249$\pm$.007$\bullet$}&\textcolor{black}{.071$\pm$.000$\circ$}&\textcolor{black}{.296$\pm$.008$\bullet$}&\textcolor{black}{\textbf{7/0/3}}\\


\textcolor{black}{LassoNet} &.015$\pm$.000$\circ$ &.020$\pm$.000$\circ$ &.019$\pm$.000$\bullet$ &.280$\pm$.011$\bullet$ &.058$\pm$.000$\bullet$ &.180$\pm$.003$\bullet$ &.023$\pm$.000$\circ$ &.119$\pm$.016$\bullet$ &.088$\pm$.001$\bullet$ &.213$\pm$.004$\bullet$ & \textbf{7/0/3} \\

\textbf{mFSIR}&.017$\pm$.000&.024$\pm$.000&.018$\pm$.000&.252$\pm$.006&.056$\pm$.001&.165$\pm$.002&.034$\pm$.000&.117$\pm$.003&.080$\pm$.000&.212$\pm$.001&-\\
  		\hline
	   & \multicolumn{11}{c} {\textbf{Ranking Loss} $\downarrow$}\\
				\cline{2-12}
    \multirow{-2}{*}{Dataset}	&bibtex&Corel16k&Delicious&emotions&Enron&language log&medical&scene&tmc2007&Yeast&\textbf{win/tie/loss counts for mFSIR}\\
		\hline
		MIFS&.218$\pm$.000$\circ$&.136$\pm$.000$\circ$&.260$\pm$.000$\circ$&.452$\pm$.001$\bullet$&.375$\pm$.000$\circ$&.170$\pm$.000$\circ$&.098$\pm$.001$\circ$&.125$\pm$.003$\bullet$&.222$\pm$.000$\circ$&.182$\pm$.007$\circ$&\textbf{2/0/8}\\
   MICO&.380$\pm$.001$\circ$&.409$\pm$.001$\circ$&.374$\pm$.001$\circ$&.311$\pm$.062$\bullet$&.368$\pm$.001$\circ$&.314$\pm$.001$\bullet$&.111$\pm$.001$\circ$&.245$\pm$.015$\bullet$&.268$\pm$.001$\circ$&.316$\pm$.018$\circ$&\textbf{3/0/7}\\
   GRRO&.470$\pm$.000$\circ$&.392$\pm$.016$\circ$&.289$\pm$.003$\circ$&.281$\pm$.006$\bullet$&.264$\pm$.005$\circ$&.362$\pm$.023$\bullet$&.081$\pm$.011$\circ$&.220$\pm$.010$\bullet$&.149$\pm$.004$\circ$&.297$\pm$.003$\circ$&\textbf{3/0/7}\\

  \textcolor{black}{CMLL}&\textcolor{black}{.206$\pm$.006$\circ$}&\textcolor{black}{.337$\pm$.012$\circ$}&\textcolor{black}{.270$\pm$.005$\circ$}&\textcolor{black}{.237$\pm$.016$\circ$}&\textcolor{black}{.288$\pm$.011$\circ$}&\textcolor{black}{.334$\pm$.013$\bullet$}&\textcolor{black}{.106$\pm$.022$\circ$}&\textcolor{black}{.135$\pm$.016$\bullet$}&\textcolor{black}{.164$\pm$.000$\circ$}&\textcolor{black}{.410$\pm$.011$\circ$}&\textcolor{black}{\textbf{2/0/8}}\\
  
  \textcolor{black}{LMLL}&\textcolor{black}{.107$\pm$.001$\circ$}&\textcolor{black}{.306$\pm$.003$\circ$}&\textcolor{black}{.608$\pm$.000$\circ$}&\textcolor{black}{.193$\pm$.018$\circ$}&\textcolor{black}{.325$\pm$.015$\circ$}&\textcolor{black}{.409$\pm$.015$\bullet$}&\textcolor{black}{.180$\pm$.017$\circ$}&\textcolor{black}{.126$\pm$.009$\bullet$}&\textcolor{black}{.098$\pm$.001$\circ$}&\textcolor{black}{.362$\pm$.010$\circ$}&\textcolor{black}{\textbf{8/0/2}}\\

  
\textcolor{black}{LassoNet} &.259$\pm$.000$\circ$ &.282$\pm$.000$\circ$ &.222$\pm$.000$\circ$ &.023$\pm$.004$\circ$ &.104$\pm$.000$\circ$ &.007$\pm$.001$\circ$ &.080$\pm$.000$\circ$ &.007$\pm$.001$\circ$ &.001$\pm$.000$\circ$ &.013$\pm$.001$\circ$ & \textbf{0/0/10} \\
  \textbf{mFSIR}&.568$\pm$.003&.756$\pm$.001&.721$\pm$.002&.271$\pm$.006&.615$\pm$.010&.312$\pm$.005&.626$\pm$.021&.105$\pm$.004&.324$\pm$.003&.433$\pm$.005&-\\
  		\hline
	   & \multicolumn{11}{c} {\textbf{Macro-averaged F1-score} $\uparrow$}\\
				\cline{2-12}
    \multirow{-2}{*}{Dataset}	&bibtex&Corel16k&Delicious&emotions&Enron&language log&medical&scene&tmc2007&Yeast&\textbf{win/tie/loss counts for mFSIR}\\
		\hline
	MIFS&.518$\pm$.000$\circ$&.034$\pm$.000$\bullet$&.602$\pm$.001$\circ$&.252$\pm$.016$\bullet$&.161$\pm$.000$\bullet$&.221$\pm$.000$\bullet$&.156$\pm$.002$\bullet$&.673$\pm$.005$\bullet$&.436$\pm$.000$\bullet$&.596$\pm$.001$\circ$&\textbf{7/0/3}\\
   MICO&.036$\pm$.018$\bullet$&.011$\pm$.001$\bullet$&.063$\pm$.000$\bullet$&.536$\pm$.067$\circ$&.135$\pm$.001$\bullet$&.115$\pm$.005$\bullet$&.313$\pm$.055$\circ$&.813$\pm$.001$\circ$&.149$\pm$.000$\bullet$&.319$\pm$.046$\bullet$&\textbf{7/0/3}\\
   GRRO&.049$\pm$.002$\bullet$&.483$\pm$.021$\circ$&.611$\pm$.003$\circ$&.563$\pm$.008$\circ$&.638$\pm$.001$\circ$&.119$\pm$.002$\bullet$&.351$\pm$.005$\circ$&.795$\pm$.009$\circ$&.847$\pm$.002$\circ$&.320$\pm$.004$\bullet$&\textbf{3/0/7}\\
   
\textcolor{black}{CMLL}&\textcolor{black}{.058$\pm$.002$\bullet$}&\textcolor{black}{.047$\pm$.000$\bullet$}&\textcolor{black}{.051$\pm$.000$\bullet$}&\textcolor{black}{.588$\pm$.022$\circ$}&\textcolor{black}{.134$\pm$.004$\bullet$}&\textcolor{black}{.439$\pm$.008$\circ$}&\textcolor{black}{.084$\pm$.004$\bullet$}&\textcolor{black}{.503$\pm$.015$\bullet$}&\textcolor{black}{.263$\pm$.000$\bullet$}&\textcolor{black}{.394$\pm$.002$\bullet$}&\textcolor{black}{\textbf{8/0/2}}\\

\textcolor{black}{LMLL}&\textcolor{black}{.060$\pm$.000$\bullet$}&\textcolor{black}{.082$\pm$.002$\bullet$}&\textcolor{black}{.037$\pm$.000$\bullet$}&\textcolor{black}{.651$\pm$.020$\circ$}&\textcolor{black}{.140$\pm$.001$\bullet$}&\textcolor{black}{.383$\pm$.006$\circ$}&\textcolor{black}{.086$\pm$.004$\bullet$}&\textcolor{black}{.543$\pm$.010$\bullet$}&\textcolor{black}{.581$\pm$.004$\circ$}&\textcolor{black}{.372$\pm$.010$\bullet$}&\textcolor{black}{\textbf{7/0/3}}\\


\textcolor{black}{LassoNet} &.131$\pm$.000$\bullet$ &.020$\pm$.000$\bullet$ &.042$\pm$.000$\bullet$ &.400$\pm$.021$\bullet$ &.123$\pm$.000$\bullet$ &.265$\pm$.012$\bullet$ &.198$\pm$.000$\bullet$ &.623$\pm$.056$\bullet$ &.241$\pm$010$\bullet$ &.391$\pm$.009$\bullet$ & \textbf{10/0/0} \\
  \textbf{mFSIR}&.156$\pm$.004&.112$\pm$.001&.099$\pm$.003&.510$\pm$.015&.383$\pm$.004&.358$\pm$.012&.292$\pm$.011&.663$\pm$.011&.497$\pm$.002&.564$\pm$.004&-\\
        \hline
	    \hline
		\label{tab:results}
	\end{tabular}
	}
\end{table*}
\subsection{Comparison on feature quality}
In this section, we present and discuss the obtained results.  Table \ref{tab:results} reports the average results (mean$\pm$std) of each comparing feature selection algorithms over ten aforementioned datasets in terms of each evaluation metric.

In addition, we also use the non-parametric \textit{Friedman test} as the statistical evaluation to analyze the relative performance among the comparing algorithms \cite{demvsar2006statistical}. Let $\gamma$  the number of algorithms, $\theta$ the number of datasets (in our case, $\gamma=7$, $\theta=10$). Accordingly, let $R_{j}=\dfrac{1}{\theta}\sum_{j}r_{i}^{j}$ denotes the average rank for the \textit{j}-th algorithm over all
datasets, with $r_{i}^{j}$  the rank of the \textit{j}-th of $\gamma$ algorithms on the \textit{i}-th of $\theta$ datasets. Then, the Friedman statistic $\mathcal{F}_{F}$ is calculated by Eq. (\ref{Eq-FF}) and is distributed according  to the \textit{F}-distribution with \textit{$\gamma$}-1 numerator degrees of freedom and (\textit{$\gamma$}-1)(\textit{$\theta$}-1) denominator degrees of freedom:

\begin{equation}
\label{Eq-FF}
\mathcal{F}_{F}=\frac{(\textit{$\theta$}-1)\mathcal{X}_{2}^{F}}{\theta (\textit{$\gamma$}-1) - \mathcal{X}_{2}^{F}}
\end{equation}

where,

\begin{equation}
\mathcal{X}_{2}^{F}=\dfrac{12\theta}{\gamma(\gamma+1)}\left[ \sum_{j=1}^{\gamma}R_{j}^{2}-\dfrac{\gamma(\gamma+1)^2}{4} \right]
\end{equation}


\begin{table}[ht]
	\centering
	\caption{Summary of the Friedman statistics $\mathcal{F}_{F}$ ($\gamma$=7 and $\theta$=10) and the critical value in terms of each evaluation metric
		($\gamma$: \# Comparing Algorithms; $\theta$: \# Datasets)}
  	\scalebox{0.8}{
	\begin{tabular}{lcc}
		\hline
		\textbf{Evaluation Metric} & \textbf{$\mathcal{F}_{F}$} & \textbf{critical value ($\alpha$=0.05)}\\
		\hline
		Hamming loss & 8.099 & \\
		Ranking loss & 7.260 & 2.84\\
		Macro-averaged F1-score & 2.189 & \\
		\hline
	\end{tabular}
	\label{FreidmanF}
 }
\end{table}

\begin{figure}
\centering
\subfloat[][Hamming loss]{
\includegraphics[width=0.35\textwidth]{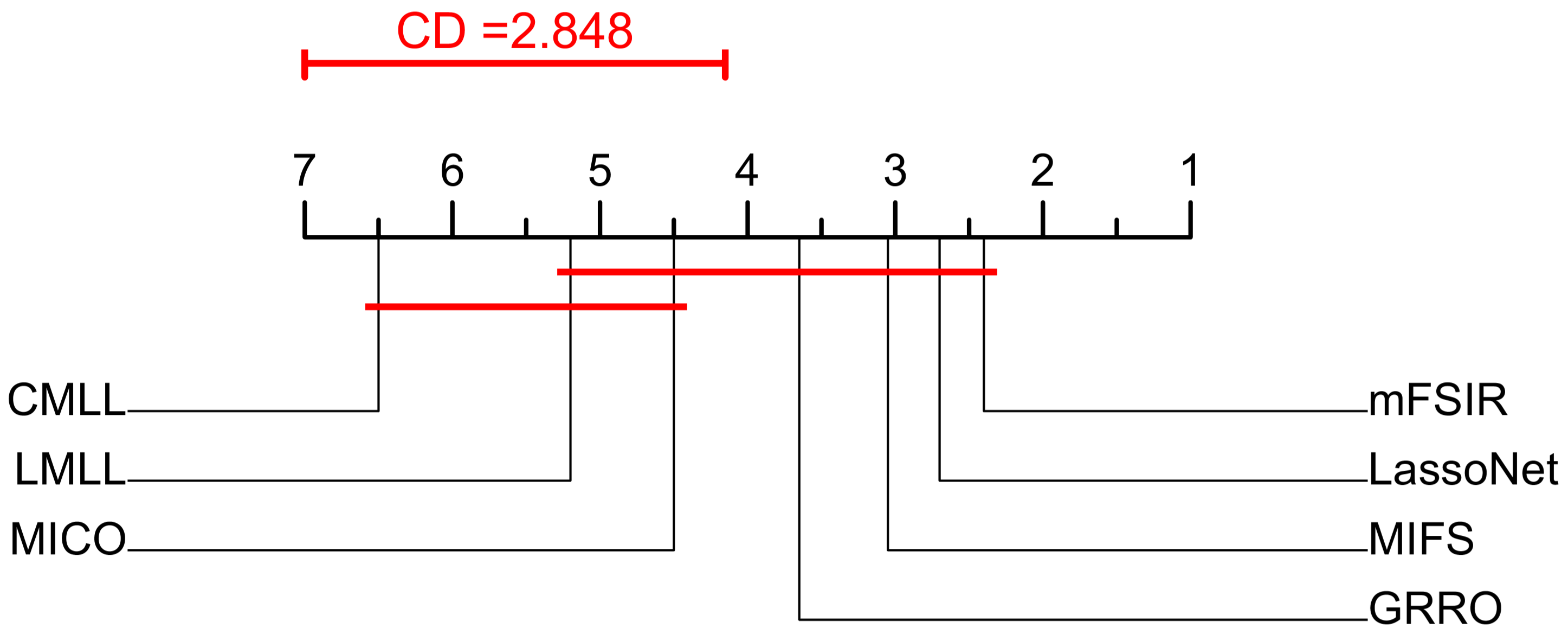}
\label{fig:subfig1}}
\qquad
\subfloat[][Ranking loss]{
\includegraphics[width=0.35\textwidth]{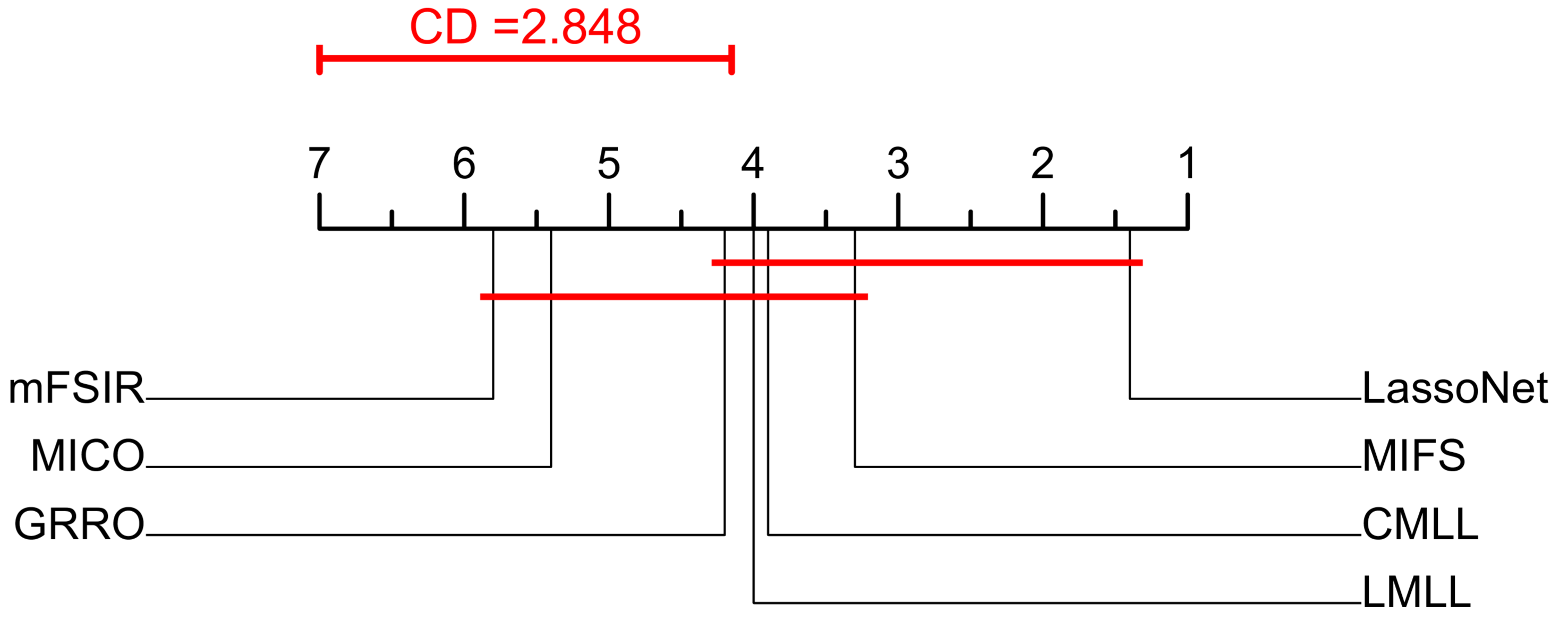}
\label{fig:subfig2}}
\qquad
\subfloat[][Macro-averaged F1-score]{
\includegraphics[width=0.35\textwidth]{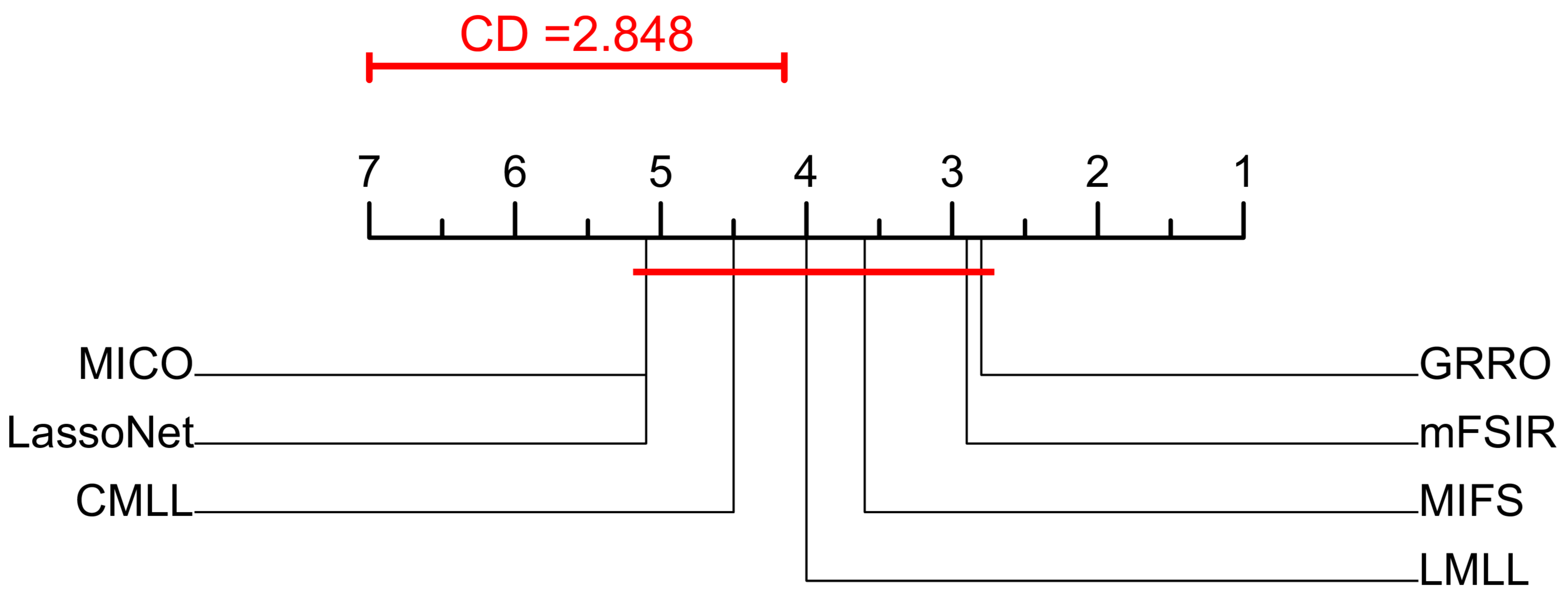}
\label{fig:subfig4}}
\caption{Comparison of mFSIR against other methods with the Nemenyi
test.}
\label{CD_freidman}
\end{figure}

Table \ref{FreidmanF} lists the  Friedman statistics $\mathcal{F}_{F}$  and the corresponding critical value in terms of each evaluation metric, at significance level $\alpha$ = 0.05. According to Table \ref{FreidmanF}, 
the null hypothesis of equal performance among the comparing approaches is clearly rejected ($\mathcal{F}_{F}$ value is greater than the critical value \textbf{2.84} in terms of all evaluation metrics). Therefore, we use the \textit{Nemenyi test} as a post-hoc test, to perform a pairwise comparison between the algorithms. The performance of two compared algorithms is deemed to be significantly  different if the difference between their corresponding average ranks is larger than or equal to at least one critical distance (CD), which is calculated by Eq. (\ref{Eq-9}).

\begin{equation}
\label{Eq-9}
CD= q_{\alpha}\sqrt{\dfrac{\gamma(\gamma+1)}{6\theta}}.
\end{equation}

where, $q_{\alpha}$ is the critical value based
on the studentized range statistics divided by $\sqrt{2}$ \cite{demvsar2006statistical}. Fig. \ref{CD_freidman} shows the CD diagrams on each evaluation metric. 
Any comparing algorithm whose average rank is within one CD to that of mFSIR is interconnected to each other with a thick line (in our case, CD =\textbf{2.84}). Otherwise, any algorithm not connected to mFSIR is considered to have significantly different performance from each other. The major observations resulting from the analysis of these results are as follows: 
\begin{itemize}
    \item As shown in Table \ref{tab:results}, and according to the overall win/tie/loss counts across all datasets at the last column of each sub-table, our mFSIR is highly competitive with other comparison algorithms in terms of all evaluation metrics. To be specific, in terms of hamming loss, mFSIR outperforms the comparing algorithms in, 70\% (LassoNet), \textcolor{black}{70\% (LMLL), 100\% (CMLL)}, 90\% (MICO), 70\% (GRRO), and 50\% (MIFS) cases. In terms of ranking loss, mFSIR outperforms the comparing algorithms in, \textcolor{black}{80\% (LMLL), 20\% (CMLL)}, 30\% (MICO), 30\% (GRRO), and 20\% (MIFS) cases. In terms of Macro-averaged F1-score, mFSIR outperforms the comparing algorithms in, 70\% (LassoNet), \textcolor{black}{70\% (LMLL), 80\% (CMLL)}, 70\% (MICO), 30\% (GRRO), and 70\% (MIFS) cases. These results mean that the implicit regularization with latent semantic  is conducive to the performance improvement. 
    \textcolor{black}{The observed performance differences between GRRO and mFSIR arise from their distinct and complementary design objectives rather than from an inherent advantage of one method over the other. GRRO is formulated to jointly optimize feature relevance and redundancy, promoting compact and globally coherent representations that capture shared structure across labels. This property is particularly beneficial in multi-label learning scenarios where controlling correlated information is essential for maintaining stable global decision functions and balanced predictive behavior across labels with heterogeneous frequencies. In contrast, mFSIR is specifically designed to preserve rich, label-wise discriminative information by optimizing feature relevance independently for each label. This design constitutes a key strength of the method, enabling precise modeling of label-specific decision boundaries and yielding strong local predictive performance when discriminative cues vary across labels. Such label-focused feature selection remains highly effective and complementary to redundancy-aware approaches. The empirical differences observed across evaluation criteria therefore reflect this methodological complementarity: GRRO emphasizes global coherence and redundancy control, whereas mFSIR prioritizes fine-grained label-level discrimination. Both perspectives address essential yet distinct challenges in multi-label learning, and their respective strengths are manifested in different aspects of predictive performance.}
	\item According to Fig. \ref{CD_freidman}, mFSIR tops the ranking in terms of hamming loss. In addition, most comparison approaches achieve statistically comparable performance.
\end{itemize}

While Table \ref{tab:results} demonstrates that mFSIR achieves comparable accuracy to existing methods on some benchmark datasets, it possesses decisive advantages for practical deployment. Most notably, its implicit regularization avoids the artificial constraints and biases inherent in explicit penalty methods (e.g., MIFS), yielding solutions that are both more flexible and theoretically robust. Additionally, as later sections of this paper will show, mFSIR is computationally significantly faster, enabling real-time applications. Furthermore, it exhibits proven stability against noise and parameter variations, ensuring reliable performance in dynamic environments.

In a nutshell, these results convincingly validate the significance of our mFSIR approach, and therefore we can safely conclude that our framework is competitive with the other well-established multi-label learning approaches.

 \subsection{Further analysis}
\subsubsection{Influence of selected features}
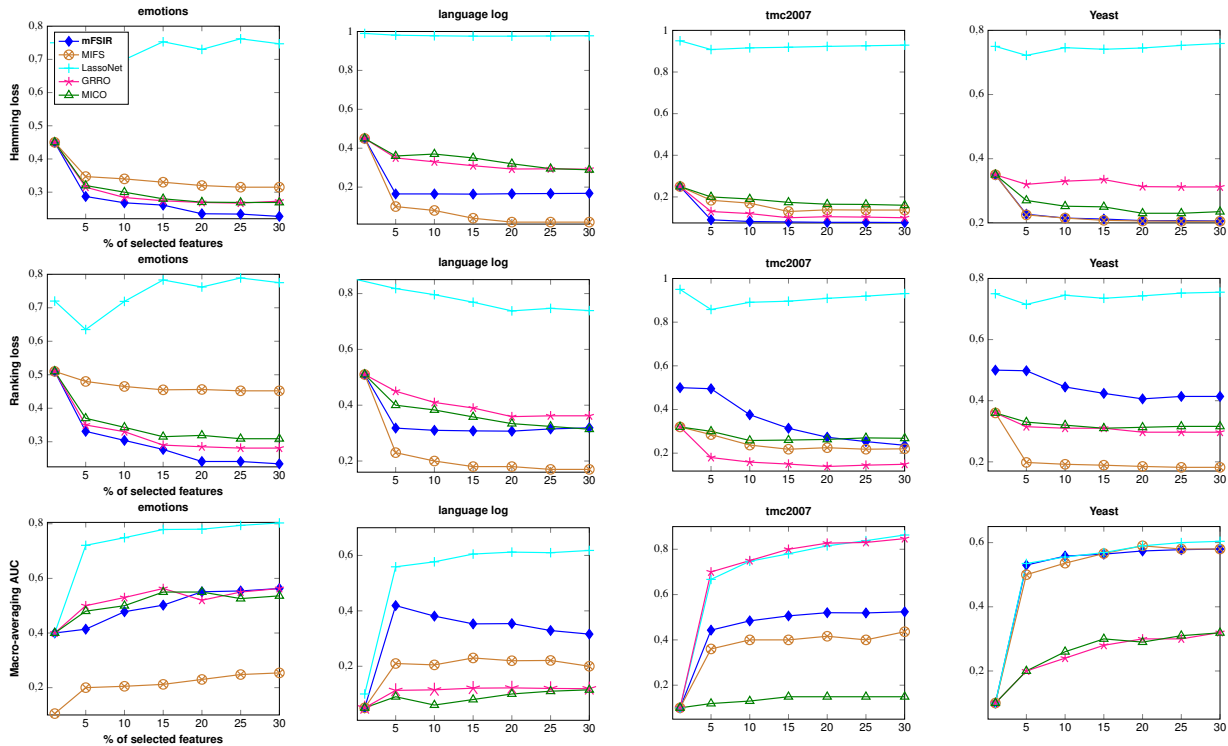
\begin{figure*}[!ht]
	\centering
	\begin{minipage}{0.22\textwidth}
		\centering
		\begin{tikzpicture} [scale=0.45]
		\begin{axis}
		[font=\sffamily,xmin=0.05,xmax=30,ymin=0.22,ymax=0.8,legend pos=north west,legend style={font=\footnotesize \sffamily },/pgf/number format/.cd,fixed,use comma,/tikz/.cd,
		xlabel=\textbf{\% of selected features},ylabel=\textbf{Hamming loss},legend cell align=left,
		xtick={0,5,10,15,20,25,30},xticklabels={0,5,10,15,20,25,30},title={\textbf{emotions}},grid style={dotted,black}]
		
		\addplot+[color=blue,mark=diamond*,mark size=4pt,mark options={color=blue},line width = 1pt] coordinates 
		{(1,0.45)(5,0.287)(10,0.268)(15,0.261)(20,0.235)(25,0.234)(30,0.227)};
		\addplot+[color=bronze, mark=otimes,mark size=4pt,mark options={color=bronze},line width = 1pt] coordinates {(1,0.45)(5,0.347)(10,0.340)(15,0.330)(20,0.320)(25,0.315)(30,0.315)};
        \addplot+[color=aqua, mark=+,mark size=4pt,solid,mark options={color=aqua},line width = 1pt] coordinates {(5,0.2927)(10,0.2549)(20,0.2429)(30,0.2241)};
        \addplot+[color=deeppink, mark=star,mark size=4pt,mark options={color=deeppink},line width = 1pt]coordinates {(5,nan)(10,nan)(20,nan)(30,0.273)};
       \addplot+[color=ao(english), mark=triangle,mark size=4pt,mark options={color=ao(english)},line width = 1pt,solid] coordinates {(5,nan)(10,nan)(20,nan)(30,0.269)};
                                    
		\legend{\textbf{mFSIR}, MIFS, LassoNet, GRRO, MICO} 
		\end{axis}	
		\end{tikzpicture}
	\end{minipage}%
	~
	\begin{minipage}{0.22\textwidth}
		\centering
		\begin{tikzpicture} [scale=0.45]
		\begin{axis}
		[font=\sffamily,xmin=0.05,xmax=30,ymin=0.01,ymax=1,legend pos=north west,legend style={font=\footnotesize \sffamily },/pgf/number format/.cd,fixed,use comma,/tikz/.cd,
		xlabel=\textbf{},ylabel=\textbf{},legend cell align=left,
		xtick={0,5,10,15,20,25,30},xticklabels={0,5,10,15,20,25,30},title={\textbf{language log}},grid style={dotted,black}]
		
		\addplot+[color=blue,mark=diamond*,mark size=4pt,mark options={color=blue},line width = 1pt] coordinates {(1,0.45)(5,0.165)(10,0.165)(15,0.164)(20,0.166)(25,0.167)(30,0.168)};
		\addplot+[color=bronze, mark=otimes,mark size=4pt,mark options={color=bronze},line width = 1pt] coordinates {(1,0.45)(5,0.1)(10,0.080)(15,0.04)(20,0.02)(25,0.02)(30,0.02)};
        \addplot+[color=aqua, mark=+,mark size=4pt,solid,mark options={color=aqua},line width=1pt] coordinates {(5,0.19479)(10,0.18580)(20,0.18342)(30,0.18023)};
        \addplot+[color=deeppink, mark=star,mark size=4pt,mark options={color=deeppink},line width = 1pt] coordinates {(5,nan)(10,nan)(20,nan)(30,0.291)};
        \addplot+[color=ao(english), mark=triangle,mark size=4pt,mark options={color=ao(english)},line width = 1pt,solid] coordinates {(5,nan)(10,nan)(20,nan)(30,0.289)};							
		\end{axis}	
		\end{tikzpicture}
	\end{minipage}%
	~
	\begin{minipage}{0.22\textwidth}
		\centering
		\begin{tikzpicture} [scale=0.45]
		\begin{axis}
		[font=\sffamily,xmin=0.05,xmax=30,ymin=0.075,ymax=1,legend pos=north west,legend style={font=\footnotesize \sffamily },/pgf/number format/.cd,fixed,use comma,/tikz/.cd,
		xlabel=\textbf{},ylabel=\textbf{},legend cell align=left,
		xtick={0,5,10,15,20,25,30},xticklabels={0,5,10,15,20,25,30},title={\textbf{tmc2007}},grid style={dotted,black}]
		
		\addplot+[color=blue,mark=diamond*,mark size=4pt,mark options={color=blue},line width = 1pt] coordinates {(1,0.25)(5,0.09)(10,0.082)(15,0.08)(20,0.078)(25,0.078)(30,0.077)};
		\addplot+[color=bronze, mark=otimes,mark size=4pt,mark options={color=bronze},line width = 1pt] coordinates {(1,0.25)(5,0.184)(10,0.17)(15,0.130)(20,0.138)(25,0.137)(30,0.137)};
        \addplot+[color=aqua, mark=+,mark size=4pt,solid,mark options={color=aqua},line width=1pt] coordinates {(5,0.09267)(10,0.09199)(20,0.08866)(30,0.08645)};
        \addplot+[color=deeppink, mark=star,mark size=4pt,mark options={color=deeppink},line width = 1pt] coordinates {(5,nan)(10,nan)(20,nan)(30,0.103)};
        \addplot+[color=ao(english), mark=triangle,mark size=4pt,mark options={color=ao(english)},line width = 1pt,solid] coordinates {(5,nan)(10,nan)(20,nan)(30,0.165)};                            
		\end{axis}	
		\end{tikzpicture}
	\end{minipage}%
	~
	\begin{minipage}{0.22\textwidth}
		\centering
		\begin{tikzpicture} [scale=0.45]
		\begin{axis}
		[font=\sffamily,xmin=0.05,xmax=30,ymin=0.200,ymax=0.8,legend pos=north west,legend style={font=\footnotesize \sffamily },/pgf/number format/.cd,fixed,use comma,/tikz/.cd,
		xlabel=\textbf{},ylabel=\textbf{},legend cell align=left,
		xtick={0,5,10,15,20,25,30},xticklabels={0,5,10,15,20,25,30},title={\textbf{Yeast}},grid style={dotted,black}]
		
		\addplot+[color=blue,mark=diamond*,mark size=4pt,mark options={color=blue},line width = 1pt] coordinates {(1,0.35)(5,0.227)(10,0.215)(15,0.212)(20,0.207)(25,0.207)(30,0.206)};
		\addplot+[color=bronze, mark=otimes,mark size=4pt,mark options={color=bronze},line width = 1pt] coordinates {(1,0.35)(5,0.225)(10,0.215)(15,0.208)(20,0.206)(25,0.205)(30,0.205)};
        \addplot+[color=aqua, mark=+,mark size=4pt,solid,mark options={color=aqua},line width=1pt] coordinates {(5,0.23839)(10,0.23244)(20,0.22064)(30,0.22035)};
        \addplot+[color=deeppink, mark=star,mark size=4pt,mark options={color=deeppink},line width = 1pt] coordinates {(5,nan)(10,nan)(20,nan)(30,0.312)};
        \addplot+[color=ao(english), mark=triangle,mark size=4pt,mark options={color=ao(english)},line width = 1pt,solid] coordinates {(5,nan)(10,nan)(20,nan)(30,0.230)};                                
		\end{axis}	
		\end{tikzpicture}
	\end{minipage}%
	~\\
	\begin{minipage}{0.22\textwidth}
		\centering
		\begin{tikzpicture} [scale=0.45]
		\begin{axis}
		[font=\sffamily,xmin=0.05,xmax=30,ymin=0.001,ymax=0.8,legend pos=north west,legend style={font=\footnotesize \sffamily },/pgf/number format/.cd,fixed,use comma,/tikz/.cd,
		xlabel=\textbf{\% of selected features},ylabel=\textbf{Ranking loss},legend cell align=left,
		xtick={0,5,10,15,20,25,30},xticklabels={0,5,10,15,20,25,30},title={\textbf{emotions}},grid style={dotted,black}]
		
		\addplot+[color=blue,mark=diamond*,mark size=4pt,mark options={color=blue},line width = 1pt] coordinates {(1,0.51)(5,0.331)(10,0.304)(15,0.277)(20,0.241)(25,0.241)(30,0.234)};
		\addplot+[color=bronze, mark=otimes,mark size=4pt,mark options={color=bronze},line width = 1pt] coordinates {(1,0.51)(5,0.480)(10,0.465)(15,0.455)(20,0.456)(25,0.452)(30,0.452)};
        \addplot+[color=aqua, mark=+,mark size=4pt,solid,mark options={color=aqua},line width = 1pt] coordinates {(5,0.02605)(10,0.02325)(20,0.01977)(30,0.01808)};
     \addplot+[color=deeppink, mark=star,mark size=4pt,mark options={color=deeppink},line width = 1pt] coordinates {(5,nan)(10,nan)(20,nan)(30,0.281)};
     \addplot+[color=ao(english), mark=triangle,mark size=4pt,mark options={color=ao(english)},line width = 1pt,solid] coordinates {(5,nan)(10,nan)(20,nan)(30,0.311)};
                                              
		\end{axis}	
		\end{tikzpicture}
	\end{minipage}%
	~
	\begin{minipage}{0.22\textwidth}
		\centering
		\begin{tikzpicture} [scale=0.45]
		\begin{axis}
		[font=\sffamily,xmin=0.05,xmax=30,ymin=0.001,ymax=0.85,legend pos=north west,legend style={font=\footnotesize \sffamily },/pgf/number format/.cd,fixed,use comma,/tikz/.cd,
		xlabel=\textbf{},ylabel=\textbf{},legend cell align=left,
		xtick={0,5,10,15,20,25,30},xticklabels={0,5,10,15,20,25,30},title={\textbf{language log}},grid style={dotted,black}]
		
		\addplot+[color=blue,mark=diamond*,mark size=4pt,mark options={color=blue},line width = 1pt] coordinates {(1,0.51)(5,0.318)(10,0.310)(15,0.308)(20,0.307)(25,0.315)(30,0.319)};
		\addplot+[color=bronze, mark=otimes,mark size=4pt,mark options={color=bronze},line width = 1pt] coordinates {(1,0.51)(5,0.230)(10,0.200)(15,0.180)(20,0.180)(25,0.170)(30,0.170)};
       \addplot+[color=aqua, mark=+,mark size=4pt,solid,mark options={color=aqua},line width=1pt]coordinates {(5,0.00537)(10,0.00788)(20,0.00796)(30,0.00809)};
        \addplot+[color=deeppink, mark=star,mark size=4pt,mark options={color=deeppink},line width = 1pt] coordinates {(5,nan)(10,nan)(20,nan)(30,0.362)};
        \addplot+[color=ao(english), mark=triangle,mark size=4pt,mark options={color=ao(english)},line width = 1pt,solid] coordinates {(5,nan)(10,nan)(20,nan)(30,0.314)};    
		\end{axis}	
		\end{tikzpicture}
	\end{minipage}%
	~
	\begin{minipage}{0.22\textwidth}
		\centering
		\begin{tikzpicture} [scale=0.45]
		\begin{axis}
		[font=\sffamily,xmin=0.05,xmax=30,ymin=0.00118,ymax=1,legend pos=north west,legend style={font=\footnotesize \sffamily },/pgf/number format/.cd,fixed,use comma,/tikz/.cd,
		xlabel=\textbf{},ylabel=\textbf{},legend cell align=left,
		xtick={0,5,10,15,20,25,30},xticklabels={0,5,10,15,20,25,30},title={\textbf{tmc2007}},grid style={dotted,black}]
		
		\addplot+[color=blue,mark=diamond*,mark size=4pt,mark options={color=blue},line width = 1pt] coordinates {(1,0.5)(5,0.495)(10,0.376)(15,0.314)(20,0.273)(25,0.253)(30,0.236)};
		\addplot+[color=bronze, mark=otimes,mark size=4pt,mark options={color=bronze},line width = 1pt] coordinates {(1,0.32)(5,0.285)(10,0.237)(15,0.218)(20,0.225)(25,0.218)(30,0.220)};
        \addplot+[color=aqua, mark=+,mark size=4pt,solid,mark options={color=aqua},line width=1pt] coordinates {(5,0.001524)(10,0.001545)(20,0.001543)(30,0.001533)};
        \addplot+[color=deeppink, mark=star,mark size=4pt,mark options={color=deeppink},line width = 1pt] coordinates {(5,nan)(10,nan)(20,nan)(30,0.149)};
        \addplot+[color=ao(english), mark=triangle,mark size=4pt,mark options={color=ao(english)},line width = 1pt,solid] coordinates {(5,nan)(10,nan)(20,nan)(30,0.268)};                            
  \end{axis}	
		\end{tikzpicture}
	\end{minipage}%
	~
	\begin{minipage}{0.22\textwidth}
		\centering
		\begin{tikzpicture} [scale=0.45]
		\begin{axis}
		[font=\sffamily,xmin=0.05,xmax=30,ymin=0.00170,ymax=0.8,legend pos=north west,legend style={font=\footnotesize \sffamily },/pgf/number format/.cd,fixed,use comma,/tikz/.cd,
		xlabel=\textbf{},ylabel=\textbf{},legend cell align=left,
		xtick={0,5,10,15,20,25,30},xticklabels={0,5,10,15,20,25,30},title={\textbf{Yeast}},grid style={dotted,black}]
		
		\addplot+[color=blue,mark=diamond*,mark size=4pt,mark options={color=blue},line width = 1pt] coordinates {(1,0.5)(5,0.498)(10,0.445)(15,0.424)(20,0.406)(25,0.414)(30,0.414)};
		\addplot+[color=bronze, mark=otimes,mark size=4pt,mark options={color=bronze},line width = 1pt] coordinates {(1,0.36)(5,0.198)(10,0.192)(15,0.189)(20,0.185)(25,0.182)(30,0.182)};
        \addplot+[color=aqua, mark=+,mark size=4pt,solid,mark options={color=aqua},line width=1pt] coordinates {(5,0.01549)(10,0.01499)(20,0.01407)(30,0.01395)};
        \addplot+[color=deeppink, mark=star,mark size=4pt,mark options={color=deeppink},line width = 1pt] coordinates {(5,nan)(10,nan)(20,nan)(30,0.297)};
         \addplot+[color=ao(english), mark=triangle,mark size=4pt,mark options={color=ao(english)},line width = 1pt,solid] coordinates {(5,nan)(10,nan)(20,nan)(30,0.316)};							
	\end{axis}	
		\end{tikzpicture}
	\end{minipage}%
	~\\
		~\\
	\begin{minipage}{0.22\textwidth}
		\centering
		\begin{tikzpicture} [scale=0.45]
		\begin{axis}
		[font=\sffamily,xmin=0.05,xmax=30,ymin=0.10,ymax=0.805,legend pos=north west,legend style={font=\footnotesize \sffamily },/pgf/number format/.cd,fixed,use comma,/tikz/.cd,
		xlabel=\textbf{\% of selected features},ylabel=\textbf{Macro-averaged F1-score},legend cell align=left,
		xtick={0,5,10,15,20,25,30},xticklabels={0,5,10,15,20,25,30},title={\textbf{emotions}},grid style={dotted,black}]
		
		\addplot+[color=blue,mark=diamond*,mark size=4pt,mark options={color=blue},line width = 1pt] coordinates {(1,0.4)(5,0.414)(10,0.478)(15,0.502)(20,0.551)(25,0.554)(30,0.563)};
		\addplot+[color=bronze, mark=otimes,mark size=4pt,mark options={color=bronze},line width = 1pt] coordinates {(1,0.105)(5,0.200)(10,0.205)(15,0.212)(20,0.23)(25,0.248)(30,0.254)};
        \addplot+[color=aqua, mark=+,mark size=4pt,solid,mark options={color=aqua},line width = 1pt] coordinates {(5,0.3783)(10,0.5352)(20,0.5608)(30,0.5592)};
        \addplot+[color=deeppink, mark=star,mark size=4pt,mark options={color=deeppink},line width = 1pt] coordinates {(5,nan)(10,nan)(20,nan)(30,0.563)};
        \addplot+[color=ao(english), mark=triangle,mark size=4pt,mark options={color=ao(english)},line width = 1pt,solid] coordinates {(5,nan)(10,nan)(20,nan)(30,0.536)};                                            
          \end{axis}	
		\end{tikzpicture}
	\end{minipage}%
	~
	\begin{minipage}{0.22\textwidth}
		\centering
		\begin{tikzpicture} [scale=0.45]
		\begin{axis}
		[font=\sffamily,xmin=0.05,xmax=30,ymin=0.005,ymax=0.7,legend pos=north west,legend style={font=\footnotesize \sffamily },/pgf/number format/.cd,fixed,use comma,/tikz/.cd,
		xlabel=\textbf{},ylabel=\textbf{},legend cell align=left,
		xtick={0,5,10,15,20,25,30},xticklabels={0,5,10,15,20,25,30},title={\textbf{language log}},grid style={dotted,black}]
		\addplot+[color=blue,mark=diamond*,mark size=4pt,mark options={color=blue},line width = 1pt] coordinates {(1,0.05)(5,0.419)(10,0.381)(15,0.353)(20,0.354)(25,0.329)(30,0.316)};
		\addplot+[color=bronze, mark=otimes,mark size=4pt,mark options={color=bronze},line width = 1pt] coordinates {(1,0.05)(5,0.210)(10,0.205)(15,0.23)(20,0.22)(25,0.221)(30,0.20)};
        \addplot+[color=aqua, mark=+,mark size=4pt,solid,mark options={color=aqua},line width=1pt]coordinates {(5,0.13440)(10,0.23894)(20,0.25229)(30,0.27022)};
        \addplot+[color=deeppink, mark=star,mark size=6pt,mark options={color=deeppink},line width = 1pt] coordinates {(5,nan)(10,nan)(20,nan)(30,0.119)};
         \addplot+[color=ao(english), mark=triangle,mark size=4pt,mark options={color=ao(english)},line width = 1pt,solid] coordinates {(5,nan)(10,nan)(20,nan)(30,0.115)};									
			\end{axis}	
		\end{tikzpicture}
	\end{minipage}%
	~
	\begin{minipage}{0.22\textwidth}
		\centering
		\begin{tikzpicture} [scale=0.45]
		\begin{axis}
		[font=\sffamily,xmin=0.05,xmax=30,ymin=0.05,ymax=0.9,legend pos=north west,legend style={font=\footnotesize \sffamily },/pgf/number format/.cd,fixed,use comma,/tikz/.cd,
		xlabel=\textbf{},ylabel=\textbf{},legend cell align=left,
		xtick={0,5,10,15,20,25,30},xticklabels={0,5,10,15,20,25,30},title={\textbf{tmc2007}},grid style={dotted,black}]
		\addplot+[color=blue,mark=diamond*,mark size=4pt,mark options={color=blue},line width = 1pt] coordinates {(1,0.1)(5,0.443)(10,0.484)(15,0.506)(20,0.52)(25,0.519)(30,0.524)};
		\addplot+[color=bronze, mark=otimes,mark size=4pt,mark options={color=bronze},line width = 1pt] coordinates {(1,0.1)(5,0.36)(10,0.40)(15,0.40)(20,0.416)(25,0.40)(30,0.436)};
        \addplot+[color=aqua, mark=+,mark size=4pt,solid,mark options={color=aqua},line width=1pt] coordinates {(5,0.09389)(10,0.15219)(20,0.22630)(30,0.28124)};
        \addplot+[color=deeppink, mark=star,mark size=4pt,mark options={color=deeppink},line width = 1pt] coordinates {(5,nan)(10,nan)(20,nan)(30,0.847)};
         \addplot+[color=ao(english), mark=triangle,mark size=4pt,mark options={color=ao(english)},line width = 1pt,solid] coordinates {(5,nan)(10,nan)(20,nan)(30,0.149)};								
		\end{axis}	
		\end{tikzpicture}
	\end{minipage}%
	~
	\begin{minipage}{0.22\textwidth}
		\centering
		\begin{tikzpicture} [scale=0.45]
		\begin{axis}
		[font=\sffamily,xmin=0.05,xmax=30,ymin=0.05,ymax=0.650,legend pos=north west,legend style={font=\footnotesize \sffamily },/pgf/number format/.cd,fixed,use comma,/tikz/.cd,
		xlabel=\textbf{},ylabel=\textbf{},legend cell align=left,
		xtick={0,5,10,15,20,25,30},xticklabels={0,5,10,15,20,25,30},title={\textbf{Yeast}},grid style={dotted,black}]
		\addplot+[color=blue,mark=diamond*,mark size=4pt,mark options={color=blue},line width = 1pt] coordinates {(1,0.1)(5,0.53)(10,0.558)(15,0.564)(20,0.574)(25,0.578)(30,0.58)};
		\addplot+[color=bronze, mark=otimes,mark size=4pt,mark options={color=bronze},line width = 1pt] coordinates {(1,0.1)(5,0.5)(10,0.536)(15,0.566)(20,0.59)(25,0.58)(30,0.58)};
        \addplot+[color=aqua, mark=+,mark size=4pt,solid,mark options={color=aqua},line width=1pt] coordinates {(5,0.28221)(10,0.32201)(20,0.38365)(30,0.37981)};
        \addplot+[color=deeppink, mark=star,mark size=4pt,mark options={color=deeppink},line width = 1pt] coordinates {(5,nan)(10,nan)(20,nan)(30,0.320)};
         \addplot+[color=ao(english), mark=triangle,mark size=4pt,mark options={color=ao(english)},line width = 1pt,solid] coordinates {(5,nan)(10,nan)(20,nan)(30,0.319)};						

		\end{axis}	
		\end{tikzpicture}
	\end{minipage}%
	
	\caption{Influence of selected feature number on four datasets {\ttfamily emotions}, {\ttfamily language log}, {\ttfamily tmc2007} and {\ttfamily Yeast}.} 
	\label{selected-features}
\end{figure*}

In this section, we study the impact of changing the number
of selected features on the performance of mFSIR. We
vary the number of selected features from 5\% to
30\%. Experiment is conducted on four datasets, including: {\ttfamily emotions}, {\ttfamily language log}, {\ttfamily tmc2007} and {\ttfamily Yeast}. Fig. \ref{selected-features} shows the performance comparison in terms of each evaluation metric across these aforementioned datasets. The major observations resulting from the analysis of these results are two-fold:
\begin{itemize}
    \item  The classification performance increases as the number of selected features increases, then keeps stable or even degrades when the number of selected features is large enough.
    \item Considering all the evaluation metrics, across the four datasets and with different numbers of selected features, mFSIR achieves highly competitive performance against the other multi-label feature selection methods. Thus, we can conclude that mFSIR  benefits to the performance, and is effective in practice.
\end{itemize}

\begin{figure*}
	\centering
	\begin{minipage}{0.22\textwidth}
		\centering
		\begin{tikzpicture} [scale=0.45]
		\begin{axis}
		[font=\sffamily,xmin=0.05,xmax=50,ymin=1
		,ymax=1296470.1719623
		,legend style={at={(0.65,0.4)},anchor=south west},
		xlabel=\textbf{Iterations},ylabel=\textbf{Objective function estimation},legend cell align=left,title={\textbf{ emotions}}]

		\addplot+[color=blue,mark=.,mark size=2pt,mark options={color=blue},line width = 2pt] coordinates {(1,589677.766590418
			)(2,533028.475721272
			)(3,484217.542243744
			)(4,442055.71417755
			)(5,405557.01393431
			)(6,373898.90416163
			)(7,346391.357463576
			)(8,322452.604851522
			)(9,301589.946514236
			)(10,283384.437699059
			)(11,267478.568690495
			)(12,253566.278714275
			)(13,241384.804512119
			)(14,230707.982714586
			)(15,221340.713019386
			)(16,213114.354979096
			)(17,205882.880866156
			)(18,199519.64486183
			)(19,193914.657768332
			)(20,188972.278791472
			)(21,184609.253311874
			)(22,180753.039155899
			)(23,177340.374584832
			)(24,174316.049709264
			)(25,171631.849807342
			)(26,169245.644460741
			)(27,167120.600810817
			)(28,165224.502801545
			)(29,163529.161186568
			)(30,162009.901467925
			)(31,160645.118906864
			)(32,159415.89138379
			)(33,158305.642248499
			)(34,157299.846443959
			)(35,156385.774147133
			)(36,155552.266980962
			)(37,154789.542538393
			)(38,154089.023543186
			)(39,153443.188470219
			)(40,152845.440873906
			)(41,152289.995038606
			)(42,151771.775878921
			)(43,151286.331288332
			)(44,150829.75536814
			)(45,150398.621170733
			)(46,149989.921766089
			)(47,149601.018592279
			)(48,149229.596182523
			)(49,148873.622476028
			)(50,148873.622476028
			)};
			
			\addplot+[color=ao(english),mark=.,mark size=2pt,mark options={color=ao(english)},line width = 2pt] coordinates {(1,1296472.1719623
			)(2,1111616.8751283
			)(3,1047319.9527043
			)(4,1024784.6853991
			)(5,1016794.2009946
			)(6,1013901.0779826
			)(7,1012808.7684815
			)(8,1012355.3736976
			)(9,1012132.6519376
			)(10,1011993.7912666
			)(11,1011884.5430804
			)(12,1011788.236432
			)(13,1011696.6768493
			)(14,1011606.1552038
			)(15,1011517.465779
			)(16,1011429.614123
			)(17,1011340.9808097
			)(18,1011252.8705652
			)(19,1011164.9548621
			)(20,1011077.0541801
			)(21,1010990.0514623
			)(22,1010902.8570936
			)(23,1010815.3944624
			)(24,1010728.5938847
			)(25,1010642.3899857
			)(26,1010555.7205318
			)(27,1010469.525884
			)(28,1010382.7487187
			)(29,1010296.3338674
			)(30,1010210.22821
			)(31,1010124.3805921
			)(32,1010039.7417551
			)(33,1009953.2642723
			)(34,1009867.9024882
			)(35,1009782.6124613
			)(36,1009697.3519079
			)(37,1009611.0801484
			)(38,1009526.7580542
			)(39,1009441.3479972
			)(40,1009356.8137992
			)(41,1009271.1206842
			)(42,1009187.2352311
			)(43,1009102.1253278
			)(44,1009017.7601268
			)(45,1008933.1100018
			)(46,1008848.1465056
			)(47,1008764.8423294
			)(48,1008679.1712628
			)(49,1008595.1081557
			)(50,1008595.1081557
			)};
			\legend{mFSIR, MIFS} 

		\end{axis}
		\end{tikzpicture}
	\end{minipage}%
	~
	\begin{minipage}{0.22\textwidth}
		\centering
		\begin{tikzpicture} [scale=0.45]
		\begin{axis}
		[font=\sffamily,xmin=0,xmax=50,ymin=1085.1081557,ymax=589677.766590418
		,legend pos=north west,legend style={font=\footnotesize \sffamily },
		xlabel=\textbf{},ylabel=\textbf{Objective function estimation},legend cell align=left,title={\textbf{ language log}}]
		
		\addplot+[color=blue,mark=.,mark size=2pt,mark options={color=blue},line width = 2pt] coordinates {(1,12964.1719623
			)(2,11116.8751283
			)(3,10473.9527043
			)(4,10247.6853991
			)(5,10167.2009946
			)(6,10139.0779826
			)(7,10128.7684815
			)(8,10123.3736976
			)(9,10121.6519376
			)(10,10119.7912666
			)(11,10118.5430804
			)(12,10117.236432
			)(13,10116.6768493
			)(14,10116.1552038
			)(15,10115.465779
			)(16,10114.614123
			)(17,10113.9808097
			)(18,10112.8705652
			)(19,10111.9548621
			)(20,10110.0541801
			)(21,10109.0514623
			)(22,10109.8570936
			)(23,10108.3944624
			)(24,10107.5938847
			)(25,10106.3899857
			)(26,10105.7205318
			)(27,10104.525884
			)(28,10103.7487187
			)(29,10102.3338674
			)(30,10102.22821
			)(31,10101.3805921
			)(32,10100.7417551
			)(33,10099.2642723
			)(34,10098.9024882
			)(35,10097.6124613
			)(36,10096.3519079
			)(37,10096.0801484
			)(38,10095.7580542
			)(39,10094.3479972
			)(40,10093.8137992
			)(41,10092.1206842
			)(42,10091.2352311
			)(43,10091.1253278
			)(44,10090.7601268
			)(45,10089.1100018
			)(46,10088.1465056
			)(47,10087.8423294
			)(48,10086.1712628
			)(49,10085.1081557
			)(50,10085.1081557
			)};
	
		\addplot+[color=ao(english),mark=.,mark size=2pt,mark options={color=ao(english)},line width = 2pt] coordinates {(1,589677.766590418
			)(2,533028.475721272
			)(3,484217.542243744
			)(4,442055.71417755
			)(5,405557.01393431
			)(6,373898.90416163
			)(7,346391.357463576
			)(8,322452.604851522
			)(9,301589.946514236
			)(10,283384.437699059
			)(11,283384.437699059
			)(12,283384.437699059
			)(13,283384.437699059
			)(14,283384.437699059
			)(15,283384.437699059
			)(16,283384.437699059
			)(17,283384.437699059
			)(18,283384.437699059
			)(19,283384.437699059
			)(20,283384.437699059
			)(21,283384.437699059
			)(22,283384.437699059
			)(23,283384.437699059
			)(24,283384.437699059
			)(25,283384.437699059
			)(26,283384.437699059
			)(27,283384.437699059
			)(28,283384.437699059
			)(29,163529.161186568
			)(30,162009.901467925
			)(31,160645.118906864
			)(32,159415.89138379
			)(33,158305.642248499
			)(34,157299.846443959
			)(35,156385.774147133
			)(36,155552.266980962
			)(37,154789.542538393
			)(38,154089.023543186
			)(39,153443.188470219
			)(40,152845.440873906
			)(41,152289.995038606
			)(42,151771.775878921
			)(43,151286.331288332
			)(44,150829.75536814
			)(45,150398.621170733
			)(46,149989.921766089
			)(47,149601.018592279
			)(48,149229.596182523
			)(49,148873.622476028
			)(50,148873.622476028
			)};
		
		\end{axis}
		\end{tikzpicture}
	\end{minipage}%
	~
	\begin{minipage}{0.22\textwidth}
		\centering
		\begin{tikzpicture} [scale=0.45]
		\begin{axis}
		[font=\sffamily,xmin=0,xmax=50,ymin=30478.065569075
		,ymax=2229830.1404241
		,legend pos=north west,legend style={font=\footnotesize \sffamily },
		xlabel=\textbf{},ylabel=\textbf{Objective function estimation},legend cell align=left,title={\textbf{tmc2007}}]
		
			\addplot+[color=blue,mark=.,mark size=2pt,mark options={color=blue},line width = 2pt] coordinates {(1,8126951.90559303
			)(2,8126951.90559303
			)(3,6169122.36179275
			)(4,6169122.36179275
			)(5,2722052.20129382
			)(6,2079776.96390833
			)(7,1592666.58595449
			)(8,1222890.15557483
			)(9,1222890.15557483
			)(10,1222890.15557483
			)(11,565391.872710621
			)(12,441281.596915534
			)(13,346527.619737796
			)(14,274085.346670361
			)(15,218618.527560581
			)(16,176081.7780908
			)(17,143405.716883173
			)(18,118259.376252341
			)(19,110000.8302596175
			)(20,109000.7030470113
			)(21,108000.7660138208
			)(22,107000.4976321609
			)(23,106000.4841445626
			)(24,106000.0475380593
			)(25,106000.622387572
			)(26,106000.2579633933
			)(27,106000.2658997547
			)(28,106000.5189344447
			)(29,106000.2719962795
			)(30,106000.6527441221
			)(31,106000.1767445687
			)(32,106000.7995160983
			)(33,106000.1362364049
			)(34,106000.5694724498
			)(35,106000.0329775986
			)(36,106000.3107785112
			)(37,106000.7294928371
			)(38,106000.1511291143
			)(39,106000.1958227834
			)(40,106000.640792797
			)(41,106000.9545716089
			)(42,106000.9352557754
			)(43,106000.4288919404
			)(44,106000.1097175969
			)(45,106000.1097175969
			)(46,106000.3082439439
			)(47,106000.8764222661
			)(48,106000.0816190923
			)(49,106000.5230242329
			)(50,106000.065569075
			)};
			\addplot+[color=ao(english),mark=.,mark size=2pt,mark options={color=ao(english)},line width = 2pt] coordinates {(1,2229830.1404241
			)(2,2200817.1708002
			)(3,2179498.2441421
			)(4,2163728.2577111
			)(5,2151997.0254748
			)(6,2143224.4041417
			)(7,2136631.8664582
			)(8,2131649.6131019
			)(9,2127862.1711873
			)(10,2124960.2076693
			)(11,2122722.6922315
			)(12,2120981.0377018
			)(13,2119604.9441467
			)(14,2118502.8964143
			)(15,2117605.573261
			)(16,2116862.9406085
			)(17,2116234.1812418
			)(18,2115693.2565924
			)(19,2115216.05
			)(20,2114789.8620789
			)(21,2114398.6198862
			)(22,2114035.5908062
			)(23,2113693.7059181
			)(24,2113367.8283623
			)(25,2113053.4725714
			)(26,2112749.6064
			)(27,2112453.2618517
			)(28,2112162.7497684
			)(29,2111876.3257325
			)(30,2111594.1931132
			)(31,2111315.7580513
			)(32,2111039.0727203
			)(33,2110765.419292
			)(34,2110493.9990584
			)(35,2110223.7001368
			)(36,2109954.923904
			)(37,2109687.4553166
			)(38,2109422.3660257
			)(39,2109158.9419963
			)(40,2108895.6294367
			)(41,2108636.9944078
			)(42,2108373.6926516
			)(43,2108114.4470562
			)(44,2107856.0308237
			)(45,2107599.2548978
			)(46,2107343.9585747
			)(47,2107089.0024897
			)(48,2106836.263379
			)(49,2106583.6301668
			)(50,2106583.6301668
			)};

		\end{axis}
		\end{tikzpicture}
	\end{minipage}%
~
	\begin{minipage}{0.22\textwidth}
		\centering
		\begin{tikzpicture} [scale=0.45]
		\begin{axis}
		[font=\sffamily,xmin=0,xmax=50,ymin=80000.1081557
		,ymax=322983.1719623
		,legend pos=north west,legend style={font=\footnotesize \sffamily },
		xlabel=\textbf{},ylabel=\textbf{Objective function estimation},legend cell align=left,title={\textbf{Yeast}}]
		
		\addplot+[color=blue,mark=.,mark size=2pt,mark options={color=blue},line width = 2pt] coordinates {(1,129647.1719623
			)(2,111161.8751283
			)(3,104731.9527043
			)(4,102478.6853991
			)(5,101679.2009946
			)(6,101390.0779826
			)(7,101280.7684815
			)(8,101235.3736976
			)(9,101213.6519376
			)(10,101199.7912666
			)(11,101188.5430804
			)(12,101178.236432
			)(13,101169.6768493
			)(14,101160.1552038
			)(15,101151.465779
			)(16,101142.614123
			)(17,101134.9808097
			)(18,101125.8705652
			)(19,101116.9548621
			)(20,101107.0541801
			)(21,101099.0514623
			)(22,101090.8570936
			)(23,101081.3944624
			)(24,101072.5938847
			)(25,101064.3899857
			)(26,101055.7205318
			)(27,101046.525884
			)(28,101038.7487187
			)(29,101029.3338674
			)(30,90000.22821
			)(31,90000.3805921
			)(32,90000.7417551
			)(33,90000.2642723
			)(34,90000.9024882
			)(35,90000.6124613
			)(36,90000.3519079
			)(37,90000.0801484
			)(38,90000.7580542
			)(39,90000.3479972
			)(40,90000.8137992
			)(41,90000.1206842
			)(42,90000.2352311
			)(43,90000.1253278
			)(44,90000.7601268
			)(45,90000.1100018
			)(46,90000.1465056
			)(47,90000.8423294
			)(48,90000.1712628
			)(49,90000.1081557
			)(50,90000.1081557
			)};
		\addplot+[color=ao(english),mark=.,mark size=2pt,mark options={color=ao(english)},line width = 2pt] coordinates {(1,340647.1719623
			)(2,211161.8751283
			)(3,204731.9527043
			)(4,202478.6853991
			)(5,101679.2009946
			)(6,101390.0779826
			)(7,101280.7684815
			)(8,101235.3736976
			)(9,101213.6519376
			)(10,101199.7912666
			)(11,101188.5430804
			)(12,101178.236432
			)(13,101169.6768493
			)(14,101160.1552038
			)(15,101151.465779
			)(16,101142.614123
			)(17,101134.9808097
			)(18,101125.8705652
			)(19,101116.9548621
			)(20,101107.0541801
			)(21,101099.0514623
			)(22,101090.8570936
			)(23,101081.3944624
			)(24,101072.5938847
			)(25,101064.3899857
			)(26,101055.7205318
			)(27,101046.525884
			)(28,101038.7487187
			)(29,101029.3338674
			)(30,101021.22821
			)(31,101012.3805921
			)(32,101003.7417551
			)(33,100995.2642723
			)(34,100986.9024882
			)(35,100978.6124613
			)(36,100969.3519079
			)(37,100961.0801484
			)(38,100952.7580542
			)(39,100944.3479972
			)(40,100935.8137992
			)(41,100927.1206842
			)(42,100918.2352311
			)(43,100910.1253278
			)(44,100901.7601268
			)(45,100893.1100018
			)(46,100884.1465056
			)(47,100876.8423294
			)(48,100867.1712628
			)(49,100859.1081557
			)(50,100859.1081557
			)};			

		\end{axis}
		\end{tikzpicture}
	\end{minipage}%
	\caption{Convergence curves of mFSIR and MIFS on four datasets {\ttfamily emotions}, {\ttfamily language log}, {\ttfamily tmc2007} and {\ttfamily Yeast}.} 
	\label{convergence}
\end{figure*}
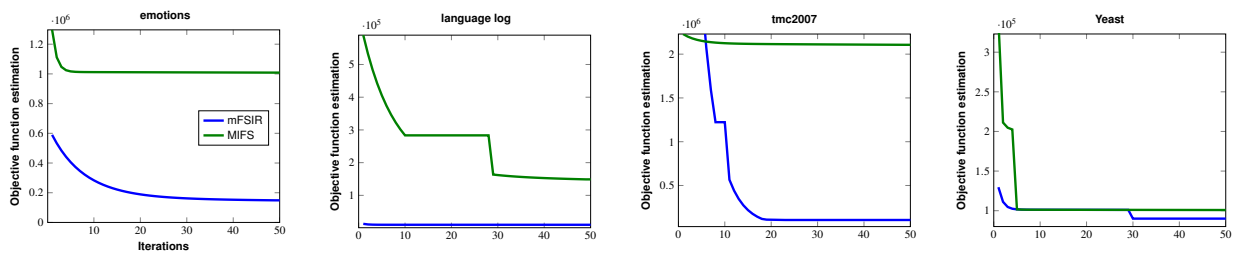
\subsubsection{ Influence of extra bias}
As explained in section \ref{related_work}, the connection between the explicit regularization term and gradient descent optimization paths can lead to a less accurate estimation. That is, the resulting estimator, by this connection, may be dominated by the bias term due to the penalty, and the estimation error cannot fall below the penalty level to accommodate a possibly faster convergence rate. To show how our proposed can overcome the extra bias and lead to more accurate estimation, we experimentally study the convergence of mFSIR which adopts the implicitly-regularized gradient descent as well as MIFS adopting the explicit term $l_{2,1}$-norm. Fig. \ref{convergence} provides an illustration of convergence curves (mFSIR and MIFS) using some datasets. The following stop criteria is used:

\begin{equation}
\frac{\zeta^t -\zeta^{t-1}}{\zeta^{t-1}} \leq 10^{-5}
\end{equation}

where $\zeta^t$ represents the objective function value in the \textit{$t^{th}$} iteration. 
As shown in Fig. \ref{convergence}, the estimation error of mFSIR is significantly lower than that of MIFS. This means that our approach  suffers significantly less from extra bias compared to MIFS and converges faster. Indeed even if MIFS ends up converging to the optimal solution, it results in a higher complexity than that of mFSIR. In contrast,  mFSIR may find a solution that optimally balances between the model complexity  and goodness fit of the model. For the four datasets {\ttfamily emotions}, {\ttfamily language log}, {\ttfamily tmc2007} and {\ttfamily Yeast}, mFSIR converges in 3 to 10 maximum iterations.

\subsubsection{Sparsity ensurement}
\label{Sparsit}
\begin{figure}[!ht]
\centering
\subfloat[][]{
\includegraphics[width=0.29\textwidth]{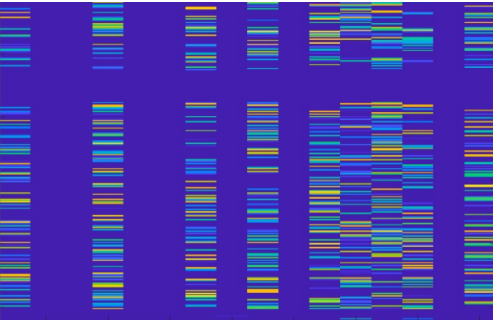}
}\\
\subfloat[][]{
\includegraphics[width=0.29\textwidth]{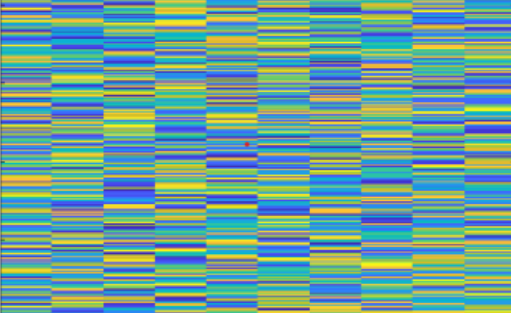}
}
\caption{\textbf{Different sparsity behaviors on {\ttfamily bibtex} dataset}. (a) represents the matrix \textbf{\^{G}}  initialized by values of $\mathbf{G}$ superior or equal to zero. The subfigure is clearly sparse with columns containing the value zero (blue color). (b) represents the matrix \textbf{\^{G}}  initialized by random values of $\mathbf{G}$. The subfigure is clearly not sparse since the columns contain different colors.}
	\label{sparsity}
\end{figure}

In this section, we show how to empirically ensure the sparsity in optimal values \textbf{\^{G}} or \textbf{\^{H}}. As mentioned in section \ref{method},  we assume  sparsity is  ensured in these two optimal values if the initial values of $\mathbf{G}$ or $\mathbf{H}$ are superior or equal to zero. Thus, we compare the  matrix of \textbf{\^{G}} or \textbf{\^{H}} after a random initialization of $\mathbf{G}$ or $\mathbf{H}$, and after an initialization with values greater than or equal to zero.
Fig. \ref{sparsity} presents the qualitative results of \textbf{\^{G}} using the {\ttfamily bibtex} dataset. From the figure, the initialization of $\mathbf{G}$ is very related to sparsity. When we initialized $\mathbf{G}$ with values greater than or equal to zero, we got a sparse matrix with columns containing the value zero (blue color) (Fig. \ref{sparsity}(a)). In contrast, when $\mathbf{G}$ initialized randomly, we got a non-sparse matrix with columns containing different colors (Fig. \ref{sparsity}(b)).

\subsubsection{Sensitivity analysis}
\begin{figure}[!ht]
	\begin{minipage}[c]{0.23\textwidth}
		\centering
	\includegraphics[width=1.2\textwidth]{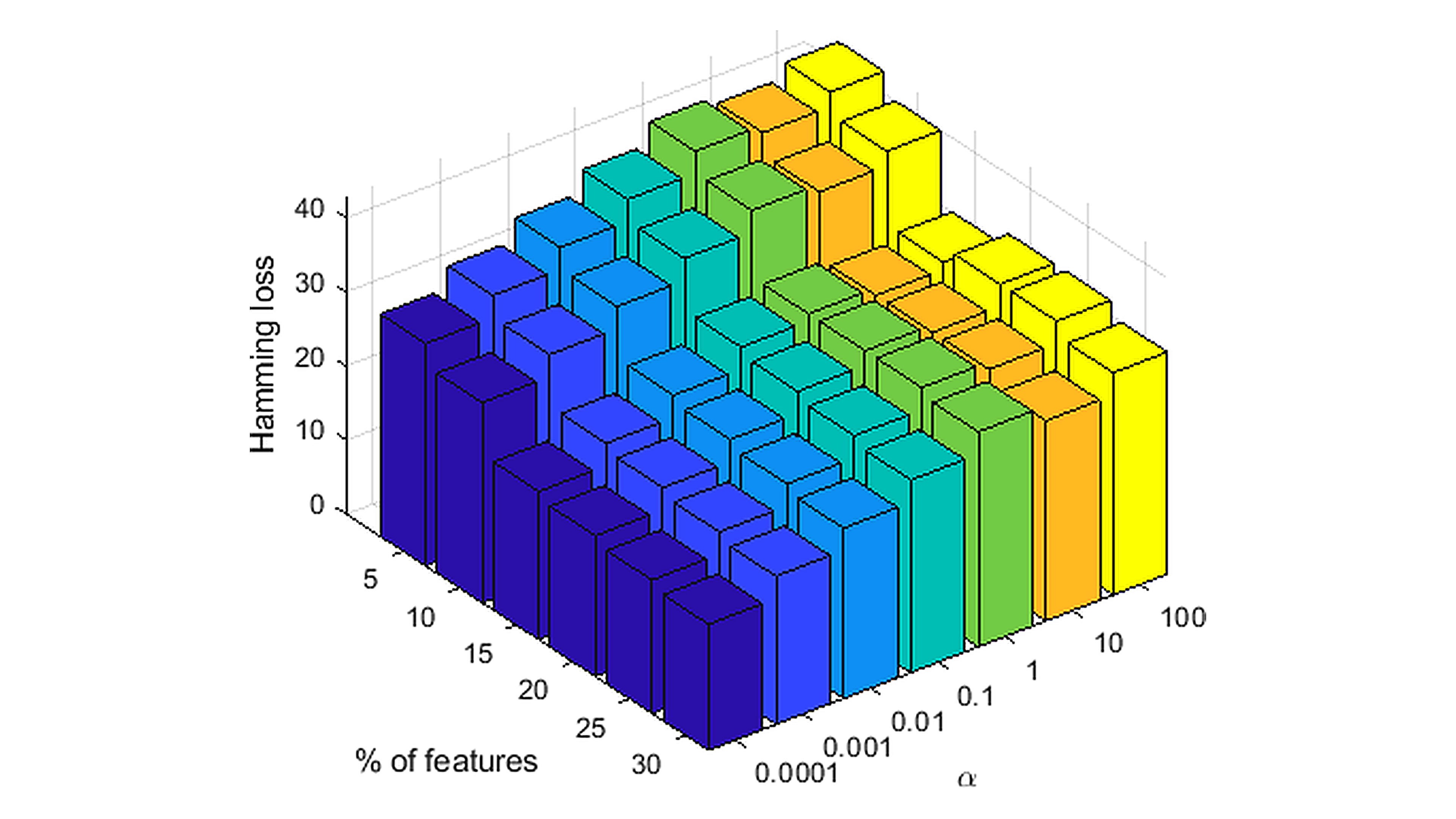}
	\end{minipage}
	\begin{minipage}[c]{0.23\textwidth}
		\centering
	\includegraphics[width=1.2\textwidth]{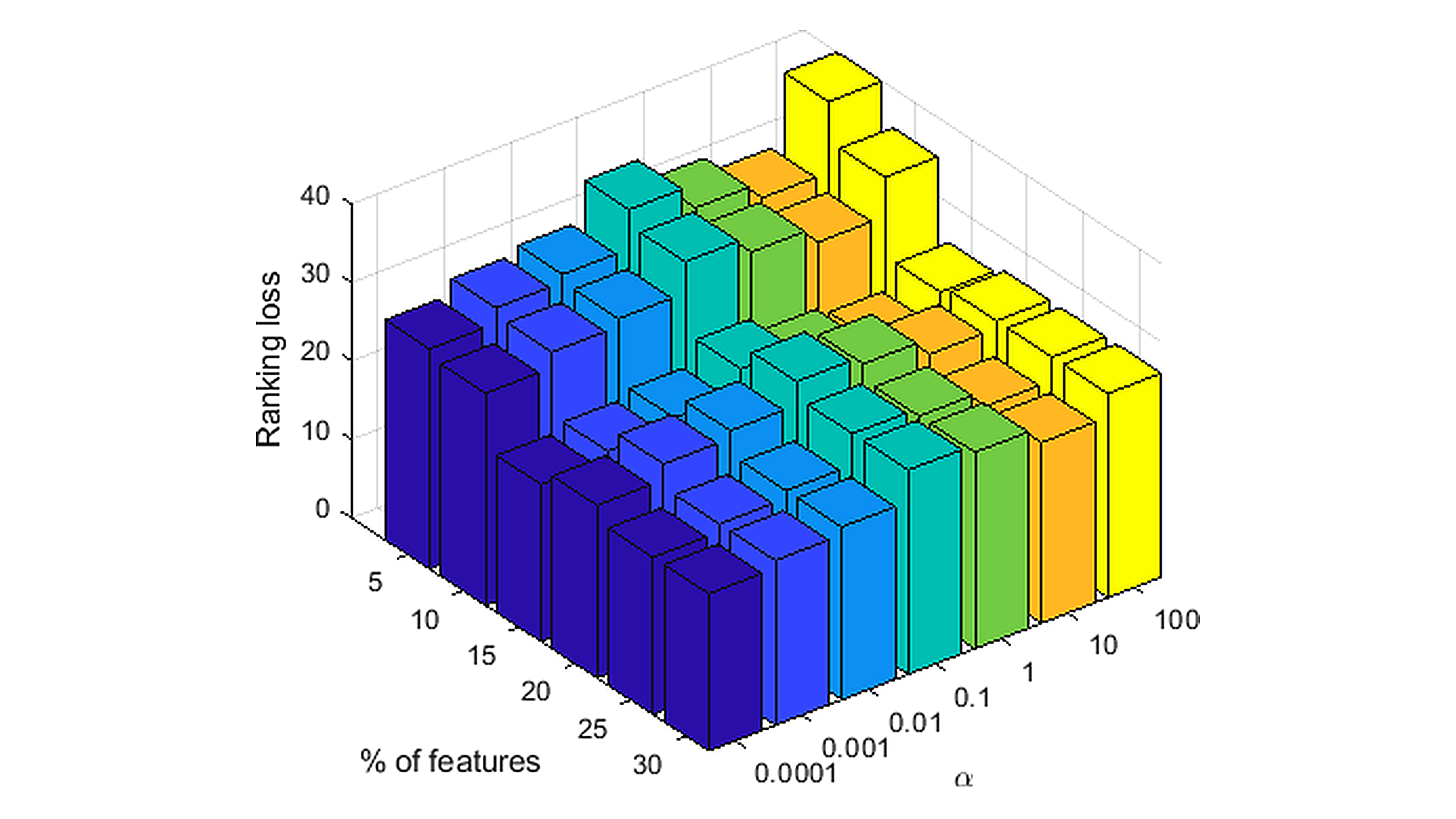}
	\end{minipage}
	\begin{minipage}[c]{0.23\textwidth}
		\centering
	\includegraphics[width=1.2\textwidth]{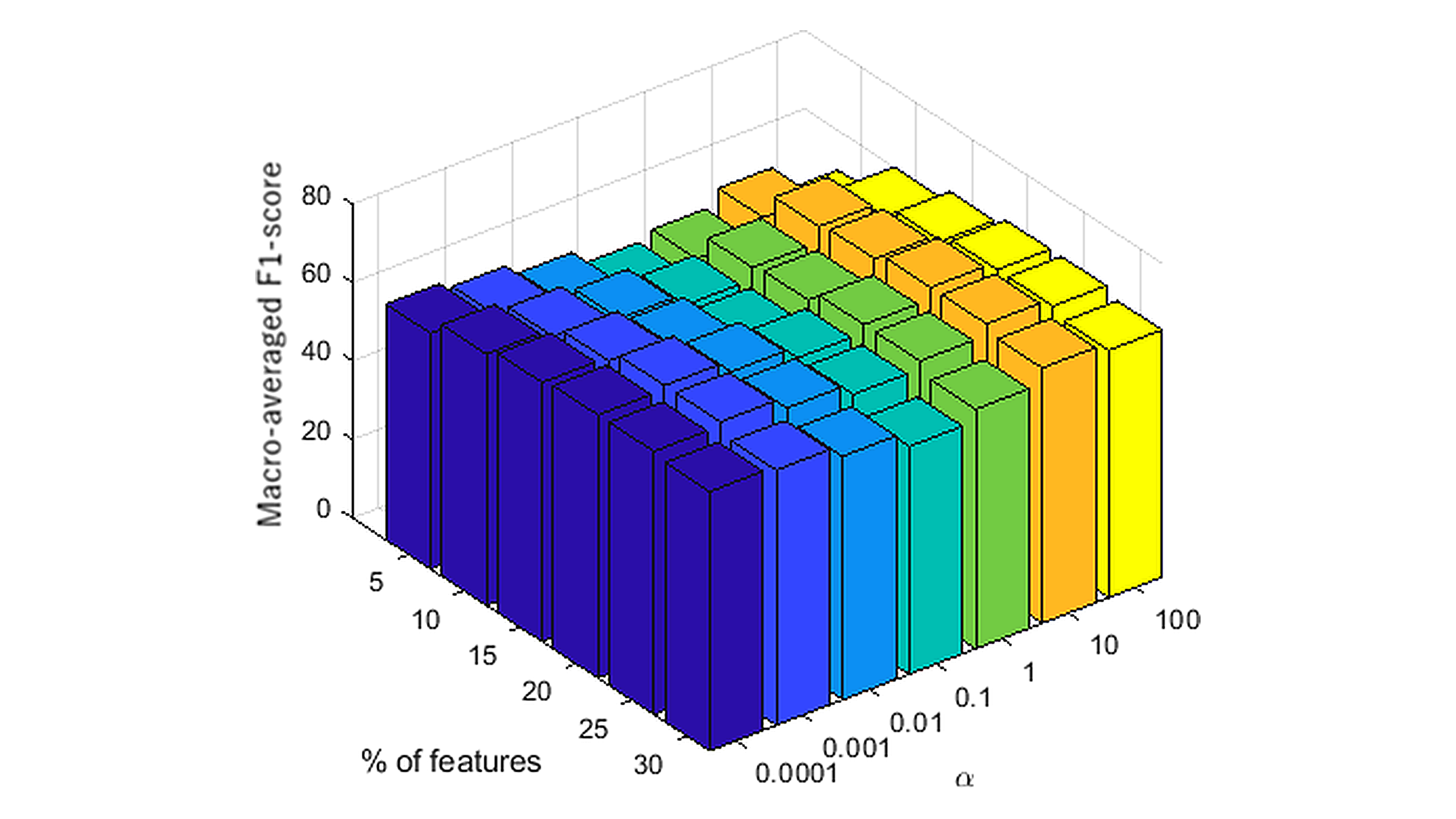}
	\end{minipage}	
	\begin{minipage}[c]{0.23\textwidth}
		\centering
	\includegraphics[width=1.2\textwidth]{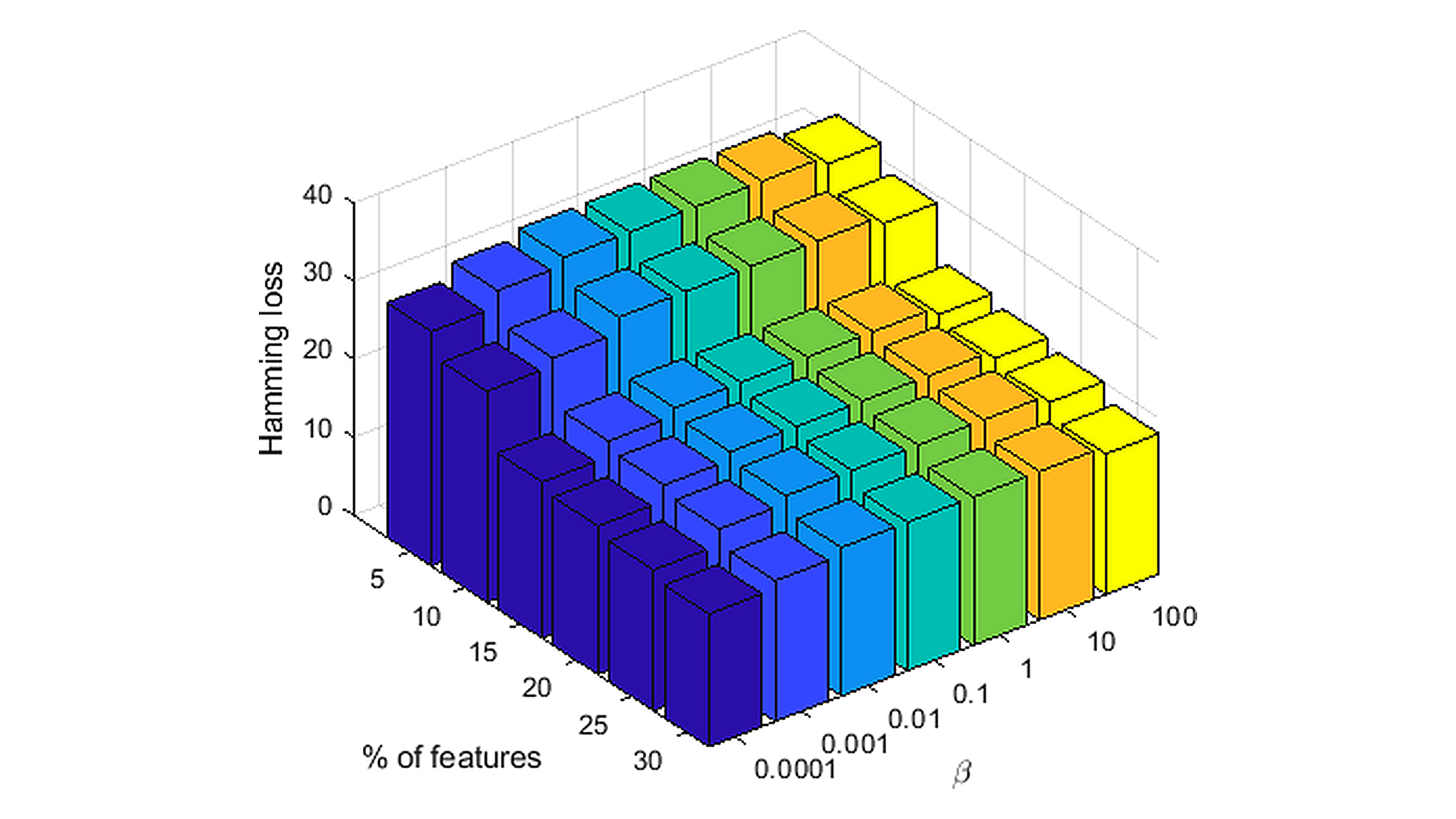}
	\end{minipage}
	\begin{minipage}[c]{0.23\textwidth}
		\centering
	\includegraphics[width=1.2\textwidth]{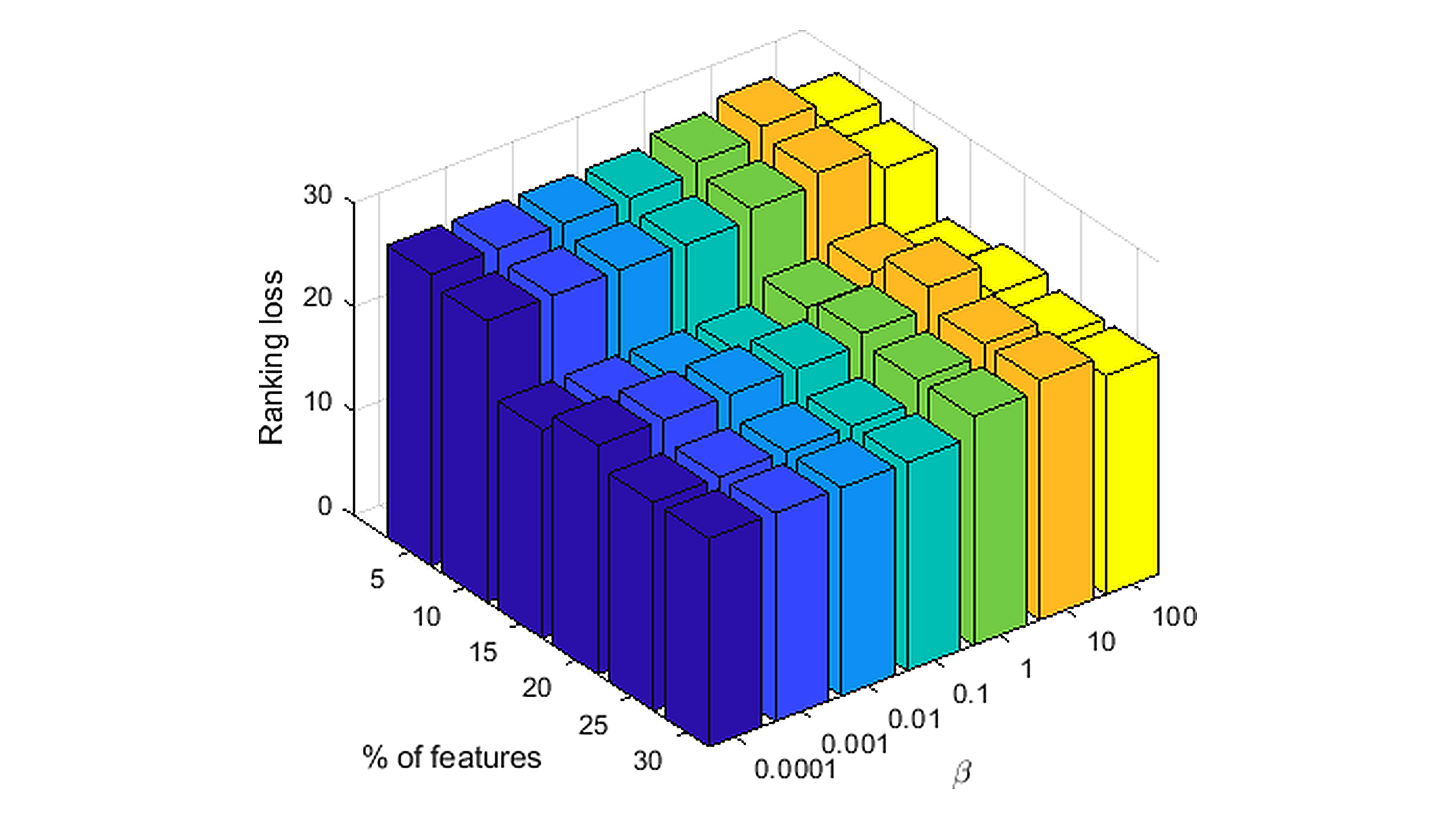}
	\end{minipage}\hspace{0.4cm}
	\begin{minipage}[c]{0.23\textwidth}
		\centering
	\includegraphics[width=1.2\textwidth]{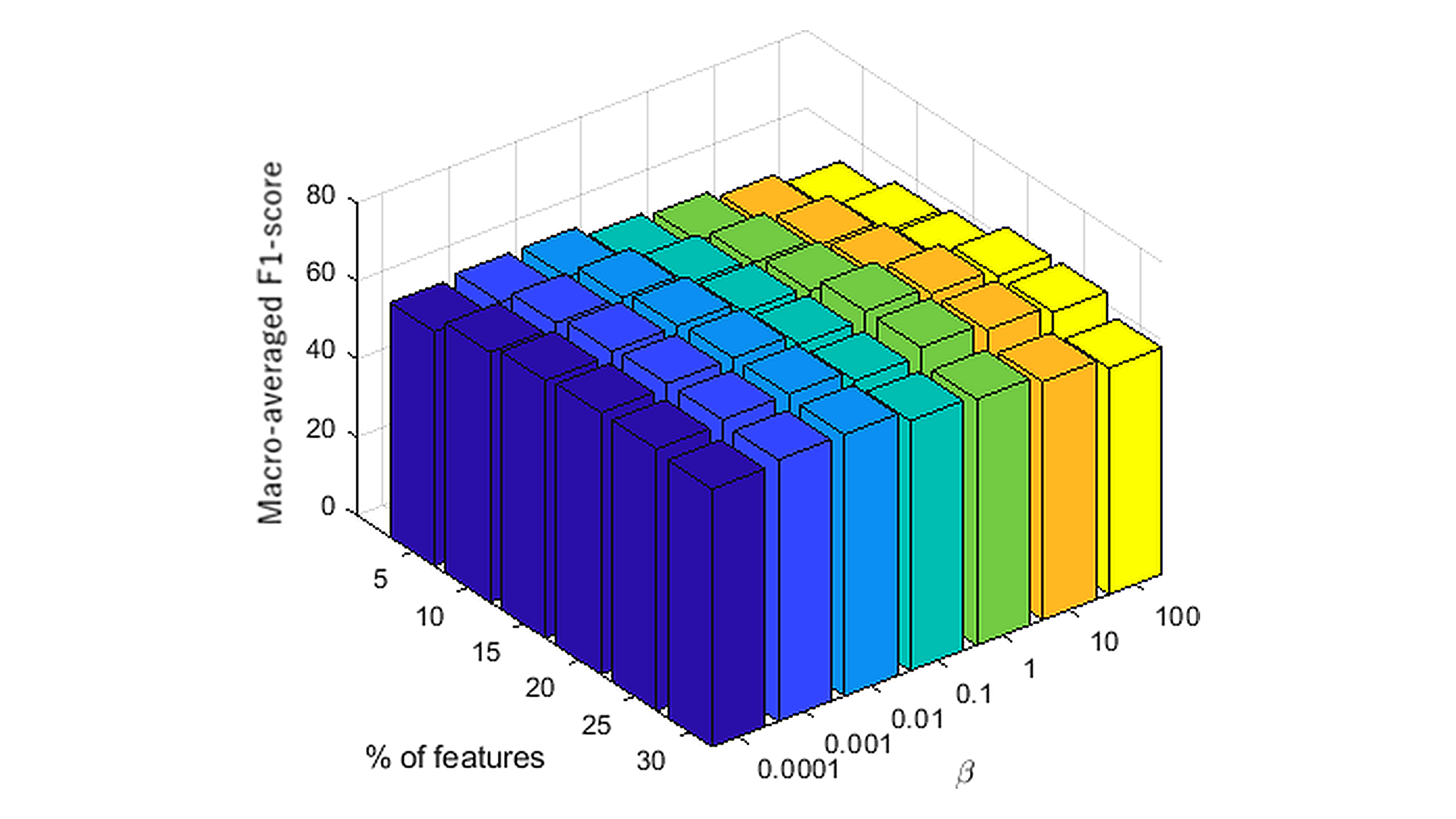}
	\end{minipage}	
 
	\caption{Performance of mFSIR changes with varying hyper-parameter configurations $\alpha$ and $\beta$ from $\{10^{-3}, 10^{-2}, 10^{-1}, 1, 10, 10^{2}\}$. Dataset: {\ttfamily emotions}; First and second row: hamming loss, ranking loss 
 and Macro-averaged F1-score v.s. regularization parameter $\alpha$ and percentage of selected features. Second and third row: hamming loss, ranking loss 
 and Macro-averaged F1-score v.s. regularization parameter $\beta$ and percentage of selected features.}
	\label{Stability parameters}
\end{figure}

In this section, we conduct an experiment to see  how the performance of mFSIR changes with varying parameter configurations $\alpha$ and $\beta$ used to balance each term in equation Eq. (\ref{OF}). We tune both parameters from $\{10^{-3}, 10^{-2}, 10^{-1}, 1, 10, 10^{2}\}$. We also vary the number of selected features from \{5\%, 10\%, 15\%, 20\%, 25\%, 30\%\} of the total number of features. Fig. \ref{Stability parameters} shows the performance of mFSIR across the {\ttfamily emotions} dataset in terms of each evaluation metric. To be specific, the first  and second  row (left subfigure) in the figure represent the performance of mFSIR in terms of hamming loss, ranking loss and Macro-averaged F1-score v.s. regularization parameter $\alpha$ ($\beta$ fixed as 1) and percentage of selected features. The second (right subfigure) and third row in the figure represent the performance of mFSIR in terms of the above three metrics v.s. regularization parameter $\beta$ ($\alpha$ fixed as 1) and percentage of selected features. From the figure, the  performance of mFSIR is not very sensitive to the changes of parameters whose values change within a reasonable range. On the other hand, the performance of mFSIR is more sensitive to the number of selected features. Additionally, with respect to all three metrics, mFSIR performance increases as the number of features selected increases.
\subsubsection{Stability analysis}
\begin{figure}[!ht]
	\centering
 	\vspace{-0.4cm}
	\includegraphics[
		width=0.35\textwidth,
		trim=0 35 0 35,
		clip
	]{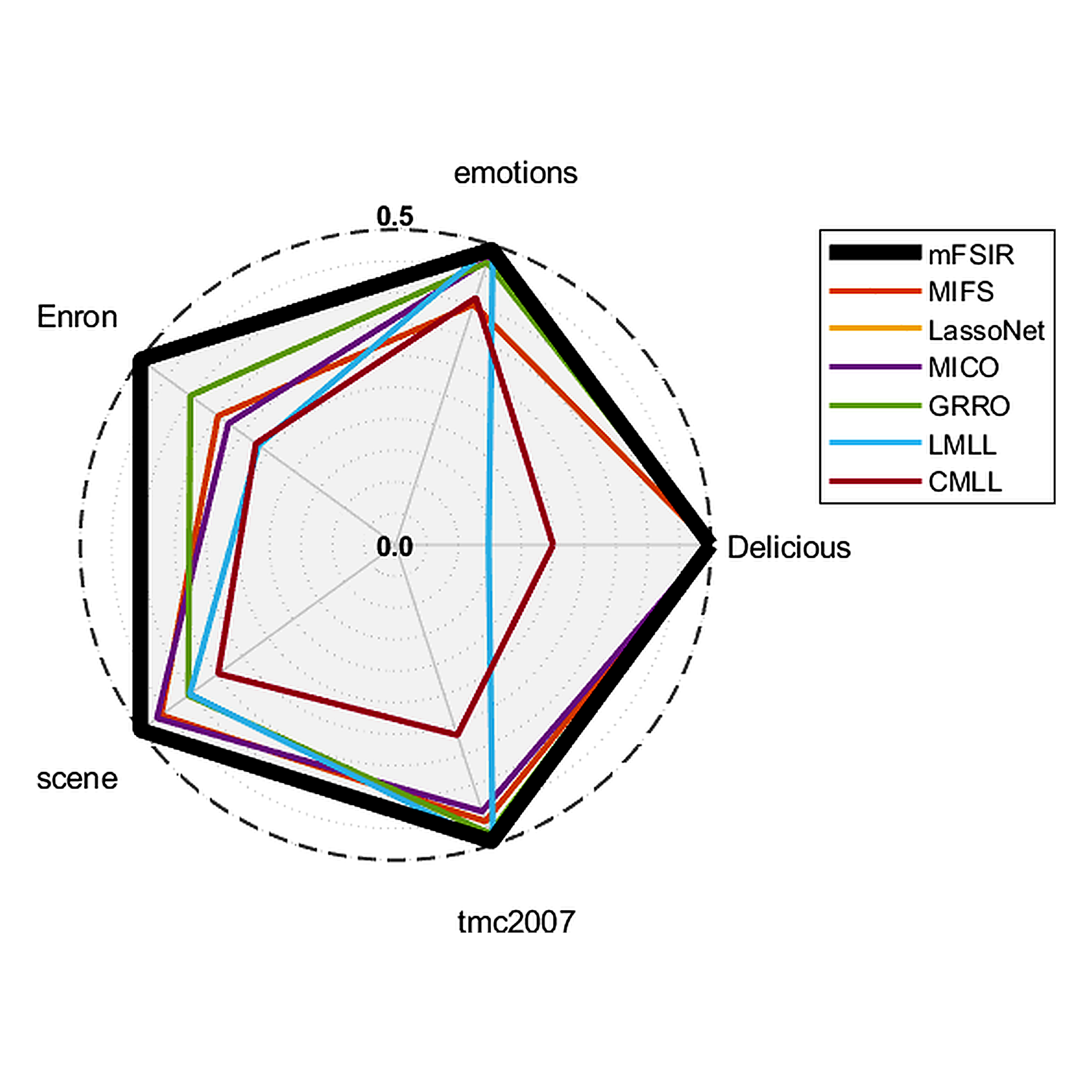}
	\vspace{-0.7cm}
	\caption{Spider Web Diagrams for stability index values considering Hamming loss metric on different multi-label datasets.}
	\label{Spider}
\end{figure}

Here, we  use the spider web diagram to examine the stability of comparing algorithms on some multi-label datasets considering hamming loss metric. The spider web graph has different corners and lines of different colors. The corners represent the different datasets and the lines represent the different algorithms. The colored area surrounded by the colored line indicates the stability value of the algorithm. The rounder and larger the area, the more stable the algorithm. Based on \cite{liu2018online}, the prediction performance is normalized into [0, 0.5]. Note that a stability value close to 0.5 is considered significant. Fig. \ref{Spider} shows the stability index according to the hamming loss values after normalization. From the figure, mFSIR is more stable compared to other algorithms, as its area almost covers the spider web graph.

\subsubsection{Runtime comparison}
\begin{figure}[!ht]
    \centering
    \includegraphics[width=0.35\textwidth]{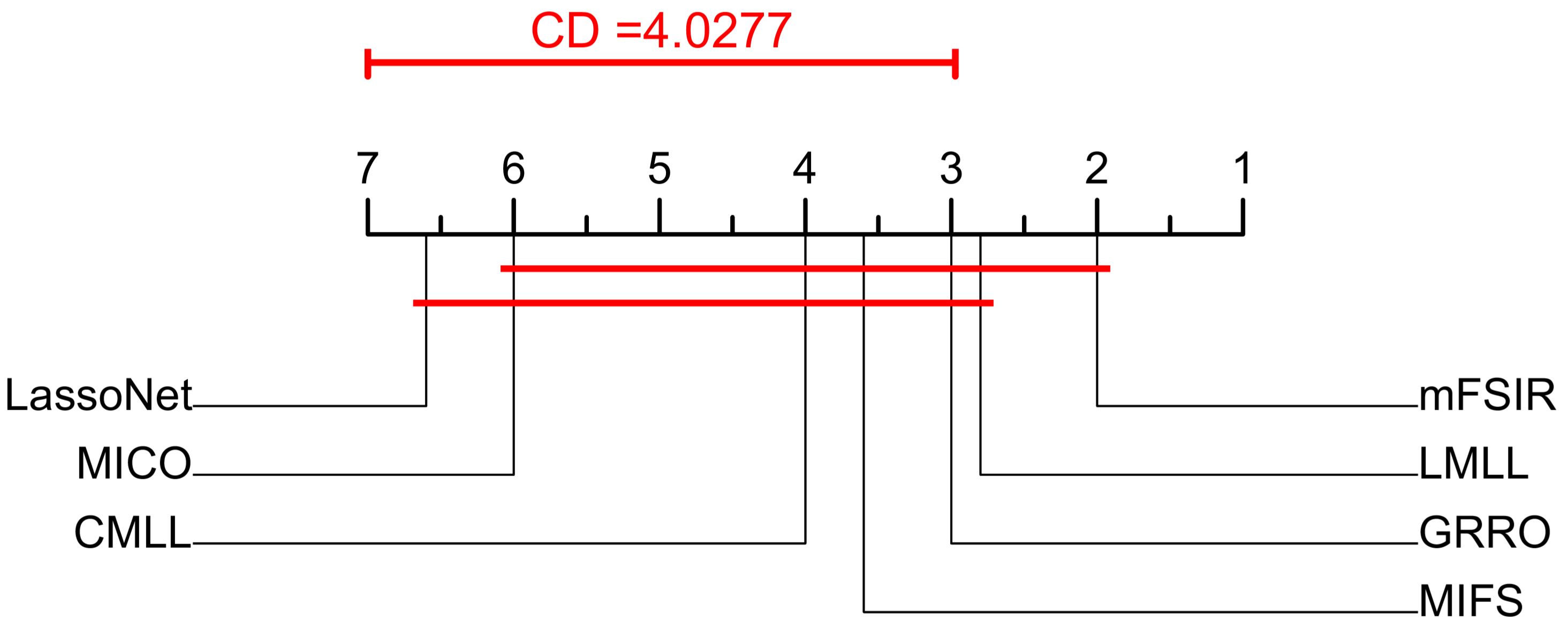}
    \caption{Performance comparison in terms of running time with the Nemenyi test.}
    \label{RunningT_CD_freidman}
\end{figure}

In this section, we illustrates the efficiency of the proposed mFSIR  by comparing its running time (in second) with the other baseline approaches on some benchmark datasets. Table \ref{Runtime} and Fig. \ref{RunningT_CD_freidman} show the average running time required to reach the convergence state of each comparison approach. 
Overall, the running time varies according to the size of the dataset. Furthermore, mFSIR is relatively comparable to the other comparing approaches and exhibits competitive runtime performance on various datasets. 

\begin{table}[!ht]
	\centering
	\caption{Performance comparison in terms of running time (sec) of different methods on some datasets. The symbole '–' indicates that time cost is over 1000 seconds.}
	\scalebox{0.7}{
	\begin{tabular}{lcccccccc}
		\hline
		\textbf{Datasets} &\textbf{MIFS} & \textbf{MICO} &\textbf{GRRO} &\textcolor{black}{CMLL}&\textcolor{black}{LMLL}&\textbf{LassoNet}&\textbf{mFSIR}\\
		\hline
		bibtex&35.55&$-$&200.61&\textcolor{black}{$-$}&\textcolor{black}{$-$}&22.71&14.22\\
		Corel16k&25.92&50.02&32.64&\textcolor{black}{$-$}&\textcolor{black}{151.79}&58.40&6.48\\
		Delicious&111.76&$-$&$-$&\textcolor{black}{$-$}&\textcolor{black}{$-$}&57.79&37.94\\
		emotions&0.61&0.64&0.05&\textcolor{black}{0.21}&\textcolor{black}{0.09}&104.54&0.15\\
		Enron&4.24&30.20&13.20&\textcolor{black}{6.24}&\textcolor{black}{18.75}&30.73&1.06\\
		language log&4.92&100.89&14.15&\textcolor{black}{2.40}&\textcolor{black}{11.82}&26.05&1.23\\
		medical&4.40&193.50&20.90&\textcolor{black}{32.96}&\textcolor{black}{0.11}&80.08&1.11\\
		scene&4.45&$-$&$-$&\textcolor{black}{11.30}&\textcolor{black}{0.57}&50.39&0.50\\
		tmc2007&51.02&$-$&$-$&\textcolor{black}{$-$}&\textcolor{black}{256.27}&97.03&17.70\\
        Yeast&1.18&0.95&0.23&\textcolor{black}{11.64}&\textcolor{black}{0.36}&36.31&0.37\\
		\hline
	\end{tabular}
	}
	\label{Runtime}
\end{table}

\subsubsection{\textcolor{black}{Ablation study}}
\textcolor{black}{In this section, we perform an ablation study on mFSIR to analyze the effects of implicit regularization and latent semantic inclusion, varying feature selection from 5\% to 30\%. 
Across {\ttfamily bibtex} and {\ttfamily medical} datasets, we compare three configurations: \textit{full mFSIR}, the original model including both implicit regularization and latent semantic inclusion; \textit{Implicit only}: mFSIR with implicit regularization only, without latent semantic inclusion; and \textit{Explicit only} (baseline), which relies solely on explicit regularization without incorporating implicit regularization or latent semantic inclusion (Eq. \ref{eqrfs}).
The results demonstrate the distinct impacts of implicit regularization and latent semantic inclusion across datasets. On {\ttfamily bibtex} (Fig. \ref{fig:combined_auc}(a)), the full mFSIR model maintains stable performance in Macro-averaged F1-score (0.141$\rightarrow$0.205), while the variant with only implicit regularization gradually declines (0.173$\rightarrow$0.117). The configuration with explicit regularization only achieves the lowest performance (0.167$\rightarrow$0.110), confirming the crucial importance of implicit regularization for stability on this dataset. In contrast, on {\ttfamily medical} (Fig. \ref{fig:combined_auc}(b)), the advantage of latent semantic inclusion becomes evident: the full model achieves the highest scores (0.568 at 30\% of features), while implicit regularization alone produces comparable results (0.571). The baseline value peaks at 0.372, demonstrating that latent semantics effectively exploits high-dimensional features on this dataset. This divergence in efficiency depending on the nature of the dataset highlights the complementarity of the two mechanisms in our approach.}

\begin{figure}[ht]
    \centering
    \subfloat[]{
        \begin{tikzpicture}
            \begin{axis}[
                legend pos=north west,
                legend style={
        at={(0.5,1.15)},
        anchor=south,
        legend columns=3,
        font=\scriptsize\sffamily,
        draw=none
    },
                xlabel=\textbf{\% of selected features},
                ylabel=\textbf{Macro-averaged F1-score},
                label style={font=\scriptsize},
    tick label style={font=\scriptsize},
                xmin=5, xmax=30,
                ymin=0.10, ymax=0.22,
                xtick={5,10,15,20,25,30},
                ytick={0.10,0.13,0.16,0.19,0.22},
                width=6.5cm,
                height=4cm
            ]
            \addplot[color=blue, mark=diamond*, mark size=3pt, very thick]
            coordinates {(5,0.141) (10,0.136) (15,0.192) (20,0.182) (25,0.185) (30,0.205)};
            \addlegendentry{Full mFSIR}

            \addplot[color=red, mark=square*, mark size=3pt, thick]
            coordinates {(5,0.133) (10,0.173) (15,0.169) (20,0.151) (25,0.143) (30,0.117)};
            \addlegendentry{Implicit only}

            \addplot[color=green!70!black, mark=triangle*, mark size=3pt, thick]
            coordinates {(5,0.122) (10,0.167) (15,0.162) (20,0.149) (25,0.120) (30,0.110)};
            \addlegendentry{Explicit only}

            \end{axis}
        \end{tikzpicture}
    }
    
    \subfloat[]{
        \begin{tikzpicture}
            \begin{axis}[
                legend pos=north west,
                legend style={
        at={(0.5,1.15)},
        anchor=south,
        legend columns=3,
        font=\scriptsize\sffamily,
        draw=none
    },
                xlabel=\textbf{\% of selected features},
                ylabel=\textbf{Macro-averaged F1-score},label style={font=\scriptsize},
    tick label style={font=\scriptsize},
                xmin=5, xmax=30,
                ymin=0.25, ymax=0.60,
                xtick={5,10,15,20,25,30},
                ytick={0.25,0.35,0.45,0.55,0.60},
                width=6.5cm,
                height=4cm
            ]
            \addplot[color=blue, mark=diamond*, mark size=3pt, very thick]
            coordinates {(5,0.316) (10,0.310) (15,0.374) (20,0.443) (25,0.572) (30,0.568)};
            \addlegendentry{Full mFSIR}

            \addplot[color=red, mark=square*, mark size=3pt, thick]
            coordinates {(5,0.314) (10,0.333) (15,0.463) (20,0.495) (25,0.488) (30,0.571)};
            \addlegendentry{Implicit only}

            \addplot[color=green!70!black, mark=triangle*, mark size=3pt, thick]
            coordinates {(5,0.266) (10,0.300) (15,0.314) (20,0.379) (25,0.361) (30,0.372)};
            \addlegendentry{Explicit only}

            \end{axis}
        \end{tikzpicture}
    }
    
  \caption{\textbf{Comparison of feature selection methods in terms of Macro-averaged F1-score.} (a) {\ttfamily bibtex} dataset. (b) {\ttfamily medical} dataset.}
    \label{fig:combined_auc}
\end{figure}


\subsubsection{Benign overfitting and implicit regularization}
\begin{figure*}[!ht]
\hspace{-2.2cm}
	\begin{minipage}[c]{0.7\textwidth}
		\centering
		\includegraphics[width=0.6\textwidth]{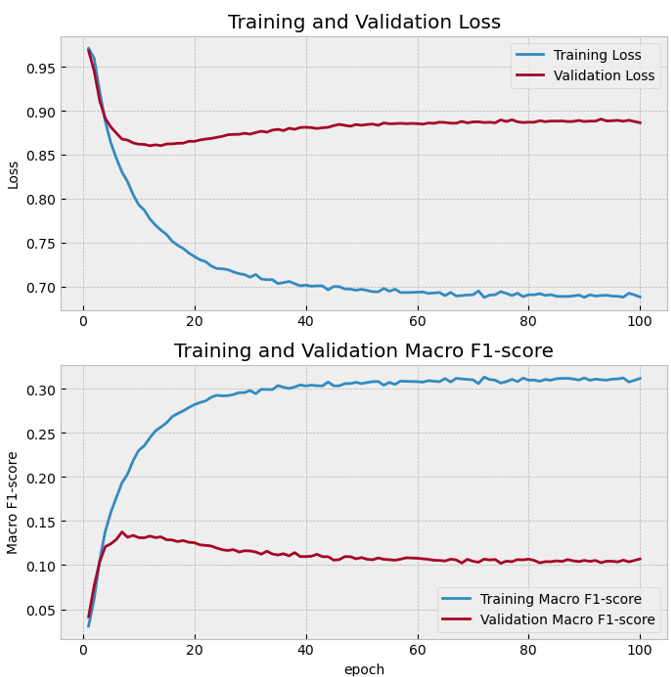}
	\end{minipage}
 \hspace{-3.5cm}
	\begin{minipage}[c]{0.7\textwidth}
		\centering
		\includegraphics[width=0.6\textwidth]{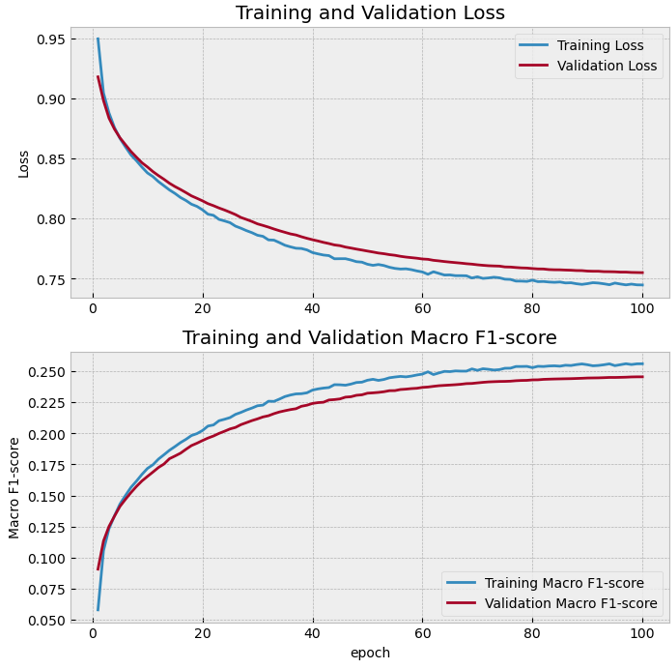}
	\end{minipage}
	\caption{\textbf{Different overfitting behaviors on {\ttfamily bibtex} dataset}. \textit{Left:} training and validation (loss and Macro-averaged F1-score) over the original {\ttfamily bibtex} (without mFSIR). \textit{Right:} training and validation (loss and Macro-averaged F1-score) over the reduced {\ttfamily bibtex} (with mFSIR, 30\% of selected features). A baseline classifier non-benignly overfits the original {\ttfamily bibtex}, while it benignly overfits the reduced {\ttfamily bibtex}.}
	\label{benign}
\end{figure*}

Here, we conduct empirical study to show that mFSIR can generalize thanks to the proposed implicit regularization and may exhibit the \textit{benign overfitting} phenomenon in some datasets. Note that this phenomenon occurs when the predictor perfectly fits the training data while achieving near optimal loss. We train a baseline classifier on the {\ttfamily bibtex} dataset for 100 epochs and test whether the baseline classifier overfits the dataset in a benign way. More precisely, we train a baseline classifier on both original {\ttfamily bibtex} dataset (without mFSIR) and the {\ttfamily bibtex} dataset processed by our mFSIR approach (the number of selected features is set at 30\%). Fig. \ref{benign} shows the training and validation loss over the aforementioned datasets. The figure also shows the training and validation Macro-averaged F1-score. The major observations resulting from the analysis of the figure are as follows:
\begin{itemize}
    \item On the reduced {\ttfamily bibtex} dataset and using the two metrics, the validation performance closely approximates the training performance (Fig. \ref{benign}, right side). However, on the original {\ttfamily bibtex} dataset, the validation performance clearly deviates from training performance (Fig. \ref{benign}, left side). This means that benign overfitting can occur on datasets cleaned of noise and unnecessary informations.
    \item Using the concept of latent space can help reduce the dataset more reliably and therefore leads to benign overfitting.
\end{itemize} 
In fact, the transition from classical overfitting (Fig. \ref{benign}, left) to benign overfitting under mFSIR (Fig. \ref{benign}, right) highlights an essential takeaway beyond a simple improved performance on the validation set: it evidences that our method performs effective feature selection by prioritizing informative structures. The observation of benign overfitting after applying mFSIR, compared to the classical overfitting observed on the original data, suggests that: (1) the training dataset constrained generalizability due to the presence of noise and irrelevant features, and (2) mFSIR improves model behaviour by capturing persistent statistical dependencies. This phenomenon corresponds to the theoretical conditions where overfitting becomes benign when the model ignores accidental correlations to focus on stable relationships \cite{bartlett2020benign}.

\subsubsection{\textcolor{black}{Comparison with modern learning models}}

\textcolor{black}{In this section, we extend the experimental analysis by comparing mFSIR with representative modern approaches for feature selection and representation learning, including LLM-based and deep architecture-based models. The objective is to position mFSIR with respect to commonly used modern baselines under a unified experimental protocol. As an LLM-based baseline, we consider \textit{all-MiniLM-L6-v2} \cite{wang2020minilm}, a compact BERT-derived model distilled from larger pre-trained language models. This model is widely used as a lightweight and efficient representation learning baseline and is particularly suitable for benchmarking under practical computational constraints. In our experiments, the representations produced by \textit{all-MiniLM-L6-v2} are evaluated using the same multi-label classifier (ML\textit{k}NN) and evaluation metrics as mFSIR. We further include DeepPINK \cite{DeepPINK18}, a deep neural network-based feature selection method, and LRDG \cite{LRDG2024}, a recent multi-label approach based on latent representation learning with dynamic graph constraints. These methods are included as representative deep and representation learning-based baselines for multi-label feature selection. All methods are evaluated under identical experimental conditions on five multi-label benchmark datasets. 
\begin{table}[!ht]
\centering
\caption{Comparison between mFSIR and modern architectures across different datasets in terms of Macro-averaged F1-score.}
\scalebox{0.8}{
\label{tab:comparison}
\begin{tabular}{lcccc}
\hline
\textbf{Dataset} &\textbf{LRDG}&\textbf{DeepPINK}& \textbf{all-MiniLM-L6-v2}  &\textbf{mFSIR} \\
\hline
Bibtex &.030 $\pm$.000&.116 $\pm$.005& \textbf{.331 $\pm$.000} & .156 $\pm$.002\\
Enron &.116 $\pm$.002&.104 $\pm$.002& .333 $\pm$.000& \textbf{.383 $\pm$.005} \\
Emotions &.462 $\pm$.000&\textbf{.547 $\pm$.031}& .521 $\pm$.027& .510 $\pm$.019 \\
Scene &.299 $\pm$.000&.568 $\pm$.029& .618 $\pm$.012& \textbf{.663 $\pm$.005} \\
Yeast &.358 $\pm$.013&.397 $\pm$.009& .332 $\pm$.001& \textbf{.564 $\pm$.004} \\
\hline
\end{tabular}
}
\end{table}
As reported in Table~\ref{tab:comparison}, mFSIR achieves competitive performance across all datasets and yields favorable results compared to the LLM-based and deep feature selection approaches on multiple datasets. These results provide a clearer positioning of mFSIR with respect to modern deep and LLM-based models.}

\textcolor{black}{Overall, the experimental results indicate that mFSIR attains competitive performance, with favorable outcomes on several datasets. This behavior can be attributed, in part, to its explicit selection of informative features, whereas representation learning-based methods generate dense embeddings optimized for general-purpose similarity, which may include features that are less relevant to the specific classification task. In contrast, mFSIR leverages implicit regularization via the Hadamard product, which naturally promotes sparsity without explicit penalization. This lightweight design can be advantageous in scenarios where computational efficiency and model transparency are important, and suggests that mFSIR can serve as a complementary approach to deep and LLM-based representation learning methods.}

\section{Conclusion and future work}
\label{conclusion}
This paper introduces a novel framework for multi-label feature selection. It is based on a new estimator that overcomes the large bias and may lead to benign overfitting. The proposed estimator relies on implicit regularization via Hadamard product parametrization in conjunction with the latent semantic analysis. Experiments validate the effectiveness of our proposed, which outperforms state-of-the-arts on several benchmark datasets. 

Our work opens up many interesting research directions, including the adaptation of implicit regularization to multi-label feature selection techniques in a semi-supervised context. 
\textcolor{black}{Another promising research direction involves extending our multi-label feature selection framework beyond unimodal data to multimodal settings. While the proposed approach focuses on unimodal multi-label data with a dedicated optimization objective, recent advances in large-scale vision–language and multimodal models highlight the potential of cross-modal learning. Exploring how embedded multi-label feature selection and implicit regularization can be combined with multimodal and alignment-based approaches represents a challenging yet promising avenue for future work.}


\bibliographystyle{IEEEtran}
\bibliography{hadamard}

@article{panos2021large,
  title={Large scale multi-label learning using Gaussian processes},
  author={Panos, Aristeidis and Dellaportas, Petros and Titsias, Michalis K},
  journal={Machine Learning},
  volume={110},
  number={5},
  pages={965--987},
  year={2021},
  publisher={Springer}
}

@article{sokolova2009systematic,
  title={A systematic analysis of performance measures for classification tasks},
  author={Sokolova, Marina and Lapalme, Guy},
  journal={Information Processing \& Management},
  year={2009}
}

@inproceedings{wei2019learning,
  title={Learning compact model for large-scale multi-label data},
  author={Wei, Tong and Li, Yu-Feng},
  booktitle={Proceedings of the AAAI Conference on Artificial Intelligence},
  volume={33},
  number={01},
  pages={5385--5392},
  year={2019}
}

@article{lv2021compact,
  title={Compact learning for multi-label classification},
  author={Lv, Jiaqi and Wu, Tianran and Peng, Chenglun and Liu, Yunpeng and Xu, Ning and Geng, Xin},
  journal={Pattern Recognition},
  volume={113},
  pages={107833},
  year={2021}
}

@article{xu2019robust,
  title={Robust multi-label learning with PRO loss},
  author={Xu, Miao and Li, Yu-Feng and Zhou, Zhi-Hua},
  journal={IEEE Transactions on Knowledge and Data Engineering},
  volume={32},
  number={8},
  pages={1610--1624},
  year={2019},
  publisher={IEEE}
}

@article{ma2018topic,
  title={Topic-based algorithm for multilabel learning with missing labels},
  author={Ma, Jianghong and Chow, Tommy WS},
  journal={IEEE transactions on neural networks and learning systems},
  volume={30},
  number={7},
  pages={2138--2152},
  year={2018},
  publisher={IEEE}
}

@article{lapin2017analysis,
  title={Analysis and optimization of loss functions for multiclass, top-k, and multilabel classification},
  author={Lapin, Maksim and Hein, Matthias and Schiele, Bernt},
  journal={IEEE transactions on pattern analysis and machine intelligence},
  volume={40},
  number={7},
  pages={1533--1554},
  year={2017},
  publisher={IEEE}
}

@article{duan2005multiple,
  title={Multiple SVM-RFE for gene selection in cancer classification with expression data},
  author={Duan, Kai-Bo and Rajapakse, Jagath C and Wang, Haiying and Azuaje, Francisco},
  journal={IEEE transactions on nanobioscience},
  volume={4},
  number={3},
  pages={228--234},
  year={2005},
  publisher={IEEE}
}

@incollection{skalak1994prototype,
  title={Prototype and feature selection by sampling and random mutation hill climbing algorithms},
  author={Skalak, David B},
  booktitle={Machine Learning Proceedings 1994},
  pages={293--301},
  year={1994},
  publisher={Elsevier}
}

@article{kohavi1997wrappers,
  title={Wrappers for feature subset selection},
  author={Kohavi, Ron and John, George H},
  journal={Artificial intelligence},
  volume={97},
  number={1-2},
  pages={273--324},
  year={1997},
  publisher={Elsevier}
}

@inproceedings{kong2012multi,
  title={Multi-label relieff and f-statistic feature selections for image annotation},
  author={Kong, Deguang and Ding, Chris and Huang, Heng and Zhao, Haifeng},
  booktitle={2012 IEEE conference on computer vision and pattern recognition},
  pages={2352--2359},
  year={2012},
  organization={IEEE}
}

@inproceedings{kononenko1994estimating,
  title={Estimating attributes: Analysis and extensions of RELIEF},
  author={Kononenko, Igor},
  booktitle={European conference on machine learning},
  pages={171--182},
  year={1994},
  organization={Springer}
}

@article{alalga20213,
  title={3-3FS: ensemble method for semi-supervised multi-label feature selection},
  author={Alalga, Abdelouahid and Benabdeslem, Khalid and Mansouri, Dou El Kefel},
  journal={Knowledge and Information Systems},
  volume={63},
  number={11},
  pages={2969--2999},
  year={2021},
  publisher={Springer}
}

@inproceedings{wu2017unified,
  title={A unified view of multi-label performance measures},
  author={Wu, Xi-Zhu and Zhou, Zhi-Hua},
  booktitle={International Conference on Machine Learning},
  pages={3780--3788},
  year={2017},
  organization={PMLR}
}

@inproceedings{li2017improving,
  title={Improving pairwise ranking for multi-label image classification},
  author={Li, Yuncheng and Song, Yale and Luo, Jiebo},
  booktitle={Proceedings of the IEEE conference on computer vision and pattern recognition},
  pages={3617--3625},
  year={2017}
}

@article{gibaja2015tutorial,
  title={A tutorial on multilabel learning},
  author={Gibaja, Eva and Ventura, Sebasti{\'a}n},
  journal={ACM Computing Surveys (CSUR)},
  volume={47},
  number={3},
  pages={1--38},
  year={2015},
  publisher={ACM New York, NY, USA}
}

@inproceedings{feng2019collaboration,
  title={Collaboration based multi-label learning},
  author={Feng, Lei and An, Bo and He, Shuo},
  booktitle={Proceedings of the AAAI Conference on Artificial Intelligence},
  volume={33},
  pages={3550--3557},
  year={2019}
}

@article{zhang2007ml,
  title={ML-KNN: A lazy learning approach to multi-label learning},
  author={Zhang, Min-Ling and Zhou, Zhi-Hua},
  journal={Pattern recognition},
  volume={40},
  number={7},
  pages={2038--2048},
  year={2007},
  publisher={Elsevier}
}

@article{sun2019mutual,
  title={Mutual information based multi-label feature selection via constrained convex optimization},
  author={Sun, Zhenqiang and Zhang, Jia and Dai, Liang and Li, Candong and Zhou, Changen and Xin, Jiliang and Li, Shaozi},
  journal={Neurocomputing},
  volume={329},
  pages={447--456},
  year={2019},
  publisher={Elsevier}
}

@inproceedings{zhang2020multiGRRO,
  title={Multi-label Feature Selection via Global Relevance and Redundancy Optimization.},
  author={Zhang, Jia and Lin, Yidong and Jiang, Min and Li, Shaozi and Tang, Yong and Tan, Kay Chen},
  booktitle={IJCAI},
  pages={2512--2518},
  year={2020}
}

@article{huang2017joint,
  title={Joint feature selection and classification for multilabel learning},
  author={Huang, Jun and Li, Guorong and Huang, Qingming and Wu, Xindong},
  journal={IEEE transactions on cybernetics},
  volume={48},
  number={3},
  pages={876--889},
  year={2017},
  publisher={IEEE}
}

@article{liu2018svm,
  title={SVM based multi-label learning with missing labels for image annotation},
  author={Liu, Yang and Wen, Kaiwen and Gao, Quanxue and Gao, Xinbo and Nie, Feiping},
  journal={Pattern Recognition},
  volume={78},
  pages={307--317},
  year={2018},
  publisher={Elsevier}
}

@article{lee2019memetic,
  title={Memetic feature selection for multilabel text categorization using label frequency difference},
  author={Lee, Jaesung and Yu, Injun and Park, Jaegyun and Kim, Dae-Won},
  journal={Information Sciences},
  volume={485},
  pages={263--280},
  year={2019},
  publisher={Elsevier}
}

@article{landauer1998introduction,
  title={An introduction to latent semantic analysis},
  author={Landauer, Thomas K and Foltz, Peter W and Laham, Darrell},
  journal={Discourse processes},
  volume={25},
  number={2-3},
  pages={259--284},
  year={1998},
  publisher={Taylor \& Francis}
}

@article{demvsar2006statistical,
  title={Statistical comparisons of classifiers over multiple data sets},
  author={Dem{\v{s}}ar, Janez},
  journal={Journal of Machine Learning Research},
  volume={7},
  pages={1--30},
  year={2006},
  publisher={JMLR. org}
}

@article{deerwester1990indexing,
  title={Indexing by latent semantic analysis},
  author={Deerwester, Scott and Dumais, Susan T and Furnas, George W and Landauer, Thomas K and Harshman, Richard},
  journal={Journal of the American society for information science},
  volume={41},
  number={6},
  pages={391--407},
  year={1990},
  publisher={Wiley Online Library}
}

@article{liu2018online,
  title={Online multi-label streaming feature selection based on neighborhood rough set},
  author={Liu, Jinghua and Lin, Yaojin and Li, Yuwen and Weng, Wei and Wu, Shunxiang},
  journal={Pattern Recognition},
  volume={84},
  pages={273--287},
  year={2018},
  publisher={Elsevier}
}

@article{jian2018exploiting,
  title={Exploiting multilabel information for noise-resilient feature selection},
  author={Jian, Ling and Li, Jundong and Liu, Huan},
  journal={ACM Transactions on Intelligent Systems and Technology (TIST)},
  volume={9},
  number={5},
  pages={1--23},
  year={2018},
  publisher={ACM New York, NY, USA}
}

@article{huang2020identification,
  title={Identification of Autistic Risk Candidate Genes and Toxic Chemicals via Multilabel Learning},
  author={Huang, Zhi-An and Zhang, Jia and Zhu, Zexuan and Wu, Edmond Q and Tan, Kay Chen},
  journal={IEEE Transactions on Neural Networks and Learning Systems},
  year={2020},
  publisher={IEEE}
}

@article{kundu2020exploiting,
  title={Exploiting weakly supervised visual patterns to learn from partial annotations},
  author={Kundu, Kaustav and Tighe, Joseph},
  journal={Advances in Neural Information Processing Systems},
  volume={33},
  year={2020}
}

@article{vito2005learning,
  title={Learning from examples as an inverse problem},
  author={Vito, Ernesto De and Rosasco, Lorenzo and Caponnetto, Andrea and Giovannini, Umberto De and Odone, Francesca},
  journal={Journal of Machine Learning Research},
  volume={6},
  number={May},
  pages={883--904},
  year={2005}
}

@article{zhao2019implicit,
  title={Implicit Regularization via Hadamard Product Over-Parametrization in High-Dimensional Linear Regression},
  author={Zhao, Peng and Yang, Yun and He, Qiao-Chu},
  journal={Biometrika},
  volume={109},
  year={2022},
  pages={1033–1046}
}

@article{zhang2019nonnegative,
  title={Nonnegative Laplacian embedding guided subspace learning for unsupervised feature selection},
  author={Zhang, Yong and Wang, Qing and Gong, Dun-wei and Song, Xian-fang},
  journal={Pattern Recognition},
  volume={93},
  pages={337--352},
  year={2019},
  publisher={Elsevier}
}

@article{song2020variable,
  title={Variable-size cooperative coevolutionary particle swarm optimization for feature selection on high-dimensional data},
  author={Song, Xian-Fang and Zhang, Yong and Guo, Yi-Nan and Sun, Xiao-Yan and Wang, Yong-Li},
  journal={IEEE Transactions on Evolutionary Computation},
  volume={24},
  number={5},
  pages={882--895},
  year={2020},
  publisher={IEEE}
}

@article{efron2004least,
  title={Least angle regression},
  author={Efron, Bradley and Hastie, Trevor and Johnstone, Iain and Tibshirani, Robert},
  journal={The Annals of statistics},
  volume={32},
  number={2},
  pages={407--499},
  year={2004},
  publisher={Institute of Mathematical Statistics}
}

@article{tibshirani1996regression,
  title={Regression shrinkage and selection via the lasso},
  author={Tibshirani, Robert},
  journal={Journal of the Royal Statistical Society: Series B (Methodological)},
  volume={58},
  number={1},
  pages={267--288},
  year={1996},
  publisher={Wiley Online Library}
}

@article{he2019joint,
  title={Joint multi-label classification and label correlations with missing labels and feature selection},
  author={He, Zhi-Fen and Yang, Ming and Gao, Yang and Liu, Hui-Dong and Yin, Yilong},
  journal={Knowledge-Based Systems},
  volume={163},
  pages={145--158},
  year={2019},
  publisher={Elsevier}
}

@article{hu2020robust,
  title={Robust multi-label feature selection with dual-graph regularization},
  author={Hu, Juncheng and Li, Yonghao and Gao, Wanfu and Zhang, Ping},
  journal={Knowledge-Based Systems},
  pages={106126},
  year={2020},
  publisher={Elsevier}
}

@article{zhang2022integrating,
  title={Integrating Global and Local Feature Selection for Multi-label Learning},
  author={Zhang, Zan and Liu, Lin and Li, Jiuyong and Wu, Xindong},
  journal={ACM Transactions on Knowledge Discovery from Data (TKDD)},
  year={2022},
  publisher={ACM New York, NY}
}

@article{paniri2020mlaco,
  title={MLACO: A multi-label feature selection algorithm based on ant colony optimization},
  author={Paniri, Mohsen and Dowlatshahi, Mohammad Bagher and Nezamabadi-Pour, Hossein},
  journal={Knowledge-Based Systems},
  volume={192},
  pages={105285},
  year={2020},
  publisher={Elsevier}
}

@book{hart2000pattern,
  title={Pattern classification},
  author={Hart, Peter E and Stork, David G and Duda, Richard O},
  year={2000},
  publisher={Wiley Hoboken}
}

@article{zhang2020multi,
  title={Multi-view Multi-label Learning with Sparse Feature Selection for Image Annotation},
  author={Zhang, Yongshan and Wu, Jia and Cai, Zhihua and Philip, S Yu},
  journal={IEEE Transactions on Multimedia},
  year={2020},
  publisher={IEEE}
}

@article{zhang2021understanding,
  title={Understanding deep learning (still) requires rethinking generalization},
  author={Zhang, Chiyuan and Bengio, Samy and Hardt, Moritz and Recht, Benjamin and Vinyals, Oriol},
  journal={Communications of the ACM},
  volume={64},
  number={3},
  pages={107--115},
  year={2021},
  publisher={ACM New York, NY, USA}
}

@article{huang2021multi,
  title={Multi-label feature selection via manifold regularization and dependence maximization},
  author={Huang, Rui and Wu, Zhejun},
  journal={Pattern Recognition},
  volume={120},
  pages={108149},
  year={2021},
  publisher={Elsevier}
}

@article{fan2021manifold,
  title={Manifold learning with structured subspace for multi-label feature selection},
  author={Fan, Yuling and Liu, Jinghua and Liu, Peizhong and Du, Yongzhao and Lan, Weiyao and Wu, Shunxiang},
  journal={Pattern Recognition},
  volume={120},
  pages={108169},
  year={2021},
  publisher={Elsevier}
}

@inproceedings{rakitianskaia2015measuring,
  title={Measuring saturation in neural networks},
  author={Rakitianskaia, Anna and Engelbrecht, Andries},
  booktitle={2015 IEEE Symposium Series on Computational Intelligence},
  pages={1423--1430},
  year={2015},
  organization={IEEE}
}

@inproceedings{vaskevicius2019implicit,
  title={Implicit regularization for optimal sparse recovery},
  author={Vaskevicius, Tomas and Kanade, Varun and Rebeschini, Patrick},
  booktitle={Advances in Neural Information Processing Systems},
  pages={2972--2983},
  year={2019}
}

@article{nie2010efficient,
  title={Efficient and robust feature selection via joint l2, 1-norms minimization},
  author={Nie, Feiping and Huang, Heng and Cai, Xiao and Ding, Chris},
  journal={Advances in neural information processing systems},
  volume={23},
  pages={1813--1821},
  year={2010}
}

@article{yao2007early,
  title={On early stopping in gradient descent learning},
  author={Yao, Yuan and Rosasco, Lorenzo and Caponnetto, Andrea},
  journal={Constructive Approximation},
  volume={26},
  number={2},
  pages={289--315},
  year={2007},
  publisher={Springer}
}

@article{dumais2004latent,
  title={Latent semantic analysis},
  author={Dumais, Susan T},
  journal={Annual review of information science and technology},
  volume={38},
  number={1},
  pages={188--230},
  year={2004},
  publisher={Wiley Online Library}
}

@inproceedings{jian2016multi,
  title={Multi-label informed feature selection.},
  author={Jian, Ling and Li, Jundong and Shu, Kai and Liu, Huan},
  booktitle={IJCAI},
  pages={1627--1633},
  year={2016}
}

@inproceedings{cai2010unsupervised,
  title={Unsupervised feature selection for multi-cluster data},
  author={Cai, Deng and Zhang, Chiyuan and He, Xiaofei},
  booktitle={Proceedings of the 16th ACM SIGKDD international conference on Knowledge discovery and data mining},
  pages={333--342},
  year={2010}
}

@inproceedings{ding2006orthogonal,
  title={Orthogonal nonnegative matrix t-factorizations for clustering},
  author={Ding, Chris and Li, Tao and Peng, Wei and Park, Haesun},
  booktitle={Proceedings of the 12th ACM SIGKDD international conference on Knowledge discovery and data mining},
  pages={126--135},
  year={2006}
}

@inproceedings{zhang2018latent,
  title={Latent semantic aware multi-view multi-label classification},
  author={Zhang, Changqing and Yu, Ziwei and Hu, Qinghua and Zhu, Pengfei and Liu, Xinwang and Wang, Xiaobo},
  booktitle={Thirty-second AAAI conference on artificial intelligence},
   pages={-},
  year={2018}
}

@inproceedings{yu2005multi,
  title={Multi-label informed latent semantic indexing},
  author={Yu, Kai and Yu, Shipeng and Tresp, Volker},
  booktitle={Proceedings of the 28th annual international ACM SIGIR conference on Research and development in information retrieval},
  pages={258--265},
  year={2005}
}

@article{shi2015sparse,
  title={Sparse feature selection based on L2, 1/2-matrix norm for web image annotation},
  author={Shi, Caijuan and Ruan, Qiuqi and Guo, Song and Tian, Yi},
  journal={Neurocomputing},
  volume={151},
  pages={424--433},
  year={2015},
  publisher={Elsevier}
}

@article{shi2019feature,
  title={Feature selection with MCP$^{2}$ regularization},
  author={Shi, Yong and Miao, Jianyu and Niu, Lingfeng},
  journal={Neural Computing and Applications},
  volume={31},
  number={10},
  pages={6699--6709},
  year={2019},
  publisher={Springer}
}

@article{fan2001variable,
  title={Variable selection via nonconcave penalized likelihood and its oracle properties},
  author={Fan, Jianqing and Li, Runze},
  journal={Journal of the American statistical Association},
  volume={96},
  number={456},
  pages={1348--1360},
  year={2001},
  publisher={Taylor \& Francis}
}

@article{hoff2017lasso,
  title={Lasso, fractional norm and structured sparse estimation using a Hadamard product parametrization},
  author={Hoff, Peter D},
  journal={Computational Statistics \& Data Analysis},
  volume={115},
  pages={186--198},
  year={2017},
  publisher={Elsevier}
}

@article{li2022benign,
  title={Benign overfitting and noisy features},
  author={Li, Zhu and Su, Weijie J and Sejdinovic, Dino},
  journal={Journal of the American Statistical Association},
  pages={1--13},
  year={2022},
  publisher={Taylor \& Francis}
}

@article{chatterji2022interplay,
  title={The interplay between implicit bias and benign overfitting in two-layer linear networks},
  author={Chatterji, Niladri S and Long, Philip M and Bartlett, Peter L},
  journal={Journal of machine learning research},
  volume={23},
  number={263},
  pages={1--48},
  year={2022}
}

@article{shamir2023implicit,
  title={The Implicit Bias of Benign Overfitting},
  author={Shamir, Ohad},
  journal={Journal of Machine Learning Research},
  volume={24},
  number={113},
  pages={1--40},
  year={2023}
}

@article{zhou2023implicit,
  title={Implicit Regularization Leads to Benign Overfitting for Sparse Linear Regression},
  author={Zhou, Mo and Ge, Rong},
  journal={arXiv preprint arXiv:2302.00257},
  year={2023}
}

@article{liu2018markov,
  title={Markov blanket and Markov boundary of multiple variables},
  author={Liu, Xu-Qing and Liu, Xin-Sheng},
  journal={The Journal of Machine Learning Research},
  volume={19},
  number={1},
  pages={1658--1707},
  year={2018},
  publisher={JMLR. org}
}

@inproceedings{liu2017deep,
  title={Deep learning for extreme multi-label text classification},
  author={Liu, Jingzhou and Chang, Wei-Cheng and Wu, Yuexin and Yang, Yiming},
  booktitle={Proceedings of the 40th International ACM SIGIR Conference on Research and Development in Information Retrieval},
  pages={115--124},
  year={2017}
}

@article{fodeh2018exploiting,
  title={Exploiting MEDLINE for gene molecular function prediction via NMF based multi-label classification},
  author={Fodeh, Samah Jamal and Tiwari, Aditya},
  journal={Journal of biomedical informatics},
  volume={86},
  pages={160--166},
  year={2018},
  publisher={Elsevier}
}

@inproceedings{wu2020multi,
  title={Multi-Label Causal Feature Selection.},
  author={Wu, Xingyu and Jiang, Bingbing and Yu, Kui and Chen, Huanhuan and Miao, Chunyan},
  booktitle={AAAI},
  pages={6430--6437},
  year={2020}
}

@article{pereira2018categorizing,
  title={Categorizing feature selection methods for multi-label classification},
  author={Pereira, Rafael B and Plastino, Alexandre and Zadrozny, Bianca and Merschmann, Luiz HC},
  journal={Artificial Intelligence Review},
  volume={49},
  number={1},
  pages={57--78},
  year={2018},
  publisher={Springer}
}

@article{zhang2010multilabel,
  title={Multilabel dimensionality reduction via dependence maximization},
  author={Zhang, Yin and Zhou, Zhi-Hua},
  journal={ACM Transactions on Knowledge Discovery from Data (TKDD)},
  volume={4},
  number={3},
  pages={1--21},
  year={2010},
  publisher={ACM New York, NY, USA}
}

@article{tsoumakas2007multi,
  title={Multi-label classification: An overview},
  author={Tsoumakas, Grigorios and Katakis, Ioannis},
  journal={International Journal of Data Warehousing and Mining (IJDWM)},
  volume={3},
  number={3},
  pages={1--13},
  year={2007},
  publisher={IGI Global}
}

@article{lassonet,
      author  = {Ismael Lemhadri and Feng Ruan and Louis Abraham and Robert Tibshirani},
      title   = {LassoNet: A Neural Network with Feature Sparsity},
      journal = {Journal of Machine Learning Research},
      year    = {2021},
      volume  = {22},
      number  = {127},
      pages   = {1-29},
      url     = {http://jmlr.org/papers/v22/20-848.html}
    }

@article{biometrica,
    author = {Zhao, Peng and Yang, Yun and He, Qiao-Chu},
    title = "{High-Dimensional Linear Regression via Implicit Regularization}",
    journal = {Biometrika},
    year = {2022},
    volume  = {109},
    number  = {4},
    pages   = {1033–1046},
    url = {https://doi.org/10.1093/biomet/asac010},
    eprint = {https://academic.oup.com/biomet/advance-article-pdf/doi/10.1093/biomet/asac010/42499850/asac010.pdf},
}

@article{wang2020minilm,
  title={Minilm: Deep self-attention distillation for task-agnostic compression of pre-trained transformers},
  author={Wang, Wenhui and Wei, Furu and Dong, Li and Bao, Hangbo and Yang, Nan and Zhou, Ming},
  journal={Advances in neural information processing systems},
  volume={33},
  pages={5776--5788},
  year={2020}
}

@article{bartlett2020benign,
  title={Benign overfitting in linear regression},
  author={Bartlett, Peter L and Long, Philip M and Lugosi, G{\'a}bor and Tsigler, Alexander},
  journal={Proceedings of the National Academy of Sciences},
  volume={117},
  number={48},
  pages={30063--30070},
  year={2020},
  publisher={National Academy of Sciences}
}

@inproceedings{DeepPINK18,
  author       = {Yang Young Lu and
                  Yingying Fan and
                  Jinchi Lv and
                  William Stafford Noble},
  title        = {DeepPINK: reproducible feature selection in deep neural networks},
  booktitle    = {Advances in Neural Information Processing Systems},
  pages        = {8690--8700},
  year         = {2018}
}

@article{LRDG2024,
title = {Multi-label feature selection via latent representation learning and dynamic graph constraints},
journal = {Pattern Recognition},
volume = {151},
pages = {110411},
year = {2024},
author = {Yao Zhang and Wei Huo and Jun Tang}
}

\end{document}